\documentclass{IEEEtaes}

\IEEEoverridecommandlockouts

\let\labelindent\relax

\usepackage{url}

\usepackage{color}
\usepackage{amsmath}
\usepackage{todonotes}
\usepackage{xspace}
\usepackage{enumitem}
\usepackage{algorithm}
\usepackage{algpseudocode}
\usepackage{amsfonts}
\usepackage{cite}
\usepackage{subcaption}
\usepackage{nameref}
\usepackage{float}
\usepackage{multirow}
\usepackage[normalem]{ulem}
\usepackage{gensymb}

\algnewcommand\Or{\textbf{or}}

\newcommand{\revone}[1]{{\color{black}{#1}}}

\newtheorem{proof}{Proof}

\graphicspath{{figs/}}

\begin{document}



\title{GATSBI: An Online GTSP-Based Algorithm for Targeted Surface Bridge Inspection and Defect Detection}

\author{Harnaik Dhami}
\member{Member, IEEE}

\author{Charith Reddy}

\author{Vishnu Dutt Sharma}
\member{Member, IEEE}

\author{Troi Williams}
\member{Member, IEEE}

\author{Pratap Tokekar}
\member{Member, IEEE}



\receiveddate{This work is supported by the National Science Foundation under grant number 1840044.}

\corresp{\itshape (Corresponding author: H. Dhami).}

\authoraddress{Dhami, Reddy, Sharma, Williams, and Tokekar are with the Department of Computer Science \& Engineering, University of Maryland, College Park, 20742, U.S.A. (email: \texttt{\small \{dhami, charith, vishnuds, troiw, tokekar\}@umd.edu}.}


\markboth{DHAMI ET AL.}{AN ONLINE ALGORITHM FOR TARGETED SURFACE BRIDGE INSPECTION AND DEFECT DETECTION}

\maketitle
\IEEEpeerreviewmaketitle

\begin{abstract}
We study the problem of visual surface inspection of infrastructure for defects using an Unmanned Aerial Vehicle (UAV). We do not assume that the geometric model of the infrastructure is known beforehand. Our planner, termed \emph{GATSBI}, plans a path in a receding horizon fashion to inspect all points on the surface of the infrastructure. The input to GATSBI consists of a 3D occupancy map created online with 3D pointclouds. Occupied voxels corresponding to the infrastructure in this map are semantically segmented and used to create an infrastructure-only occupancy map. Inspecting an infrastructure voxel requires the UAV to take images from a desired viewing angle and distance. We then create a Generalized Traveling Salesperson Problem (GTSP) instance to cluster candidate viewpoints for inspecting the infrastructure voxels and use an off-the-shelf GTSP solver to find the optimal path for the given instance. As the algorithm sees more parts of the environment over time, it replans the path to inspect uninspected parts of the infrastructure while avoiding obstacles. We evaluate the performance of our algorithm through high-fidelity simulations conducted in AirSim and real-world experiments. We compare the performance of GATSBI with a baseline inspection algorithm where the map is known a priori. Our evaluation reveals that targeting the inspection to only the segmented infrastructure voxels and planning carefully using a GTSP solver leads to a more efficient and thorough inspection than the baseline inspection algorithm.
\end{abstract}

\section{Introduction}

In this work, we are interested in designing a high-level planner for an Unmanned Aerial Vehicle (UAV) that inspects a 3D surface for identifying visual defects. Currently, infrastructure inspection is performed manually either by an inspector being suspended across the infrastructures surface or by the inspector piloting a UAV. The first is dangerous whereas the second prevents the inspector from completely focusing on detecting defects, such as cracks.  Developing an autonomous inspection planner addresses these concerns.

Recently, a number of commercial solutions such as the ones from Skydio~\cite{skydio} and Exyn~\cite{Exyn_Technologies_undated-gq} and ongoing work in academia provide robust autonomy including SLAM and low-level planning (how to navigate from point A to point B). Our work on high-level planning (determining what the next waypoint B should be) is complementary to these works. Current forms of planning mostly consist of someone clicking on waypoints for the UAVs to fly to. As a result, we develop tools that autonomously solve the more general problem of inspecting infrastructure with no prior information about its geometry.

Inspection is closely related to coverage and exploration, which are problems that have been well-studied in the literature. However, coverage and exploration are not necessarily the best approaches for inspection. Given a 3D model of the environment (including the infrastructure), we can find a coverage path that covers all points on the infrastructure using an offline planner~\cite{9048979}. In practice, we often do not have any prior model of the layout of the infrastructure. Even if a prior 3D model is available, it may be inaccurate due to changes in the environment surrounding the infrastructure as well as structural changes made to the infrastructure. In this work, we address the problem of designing targeted inspection plans as the 3D model of the environment is built online. 

\begin{figure}
    \centering
    \includegraphics[width = 0.75\columnwidth]{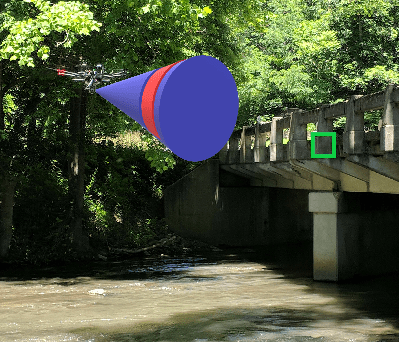}
    \caption{An example of $View$. Our goal is to ensure that each point on the surface is viewed by the onboard camera within the \emph{View} region. The green box depicts the face of an infrastructure voxel, the blue cone depicts the viewing cone, and the red band on the cone depicts the viewing distance.}
    \label{fig:vcvd}
\end{figure}

Frontier-based strategies~\cite{yamauchi1997frontier} are typically used for exploring an initially unknown environment. A frontier is a boundary between explored and unexplored regions. The strategy chooses which of the frontiers to visit (and which path to follow to get to the chosen frontier) to help speed up exploration. The algorithm terminates when there are no more accessible unexplored regions. A bounding box placed around the infrastructure can restrict exploration when operating in an open environment. Exploring the infrastructure does not necessarily mean that the UAV will get inspection-quality images. Instead, an inspection planner that can take into account viewing and distance constraints may be more efficient. We present such a planner, termed \emph{GTSP-Based Algorithm for Targeted Surface Bridge Inspection (GATSBI)}, and show that it outperforms previous inspection strategies in efficiently inspecting the infrastructure (Section~\ref{sec:eval}).

GATSBI consists of six modules: semantic segmentation to find 3D points from a pointcloud that corresponds to the surface of the infrastructure, simultaneous localization and mapping (SLAM) to align each incoming 3D pointcloud, 3D occupancy grid mapping using the 3D pointcloud, a high-level planner for finding inspection paths for the UAV, a navigation algorithm for executing the planned path, and a crack detection neural network to detect defects. We use off-the-shelf modules for SLAM (LIO-SAM~\cite{liosam2020shan}), occupancy grid mapping (OctoMap~\cite{hornung2013octomap}), solving GTSP (GLNS~\cite{Smith2016GLNS}), point-to-point navigation (MoveIt~\cite{moveit}), and crack detection (YOLO-World~\cite{Cheng_2024_CVPR}). The specific point-to-point planner we use is RRT*. Since the distances generated by RRT* might be different than the Euclidean distance, we perform a lazy evaluation prior to navigation described in Section~\ref{sec:GATSBI}. Our key algorithmic contribution is the high-level planner module. Specifically, we show how to show \emph{how} to reduce the inspection problem to a GTSP instance and the full pipeline that outperforms baseline strategies. Our technique takes into account overlapping viewpoints that can view the same parts of the infrastructure and simultaneously selects \emph{where} to take images and \emph{what order} to visit those viewpoints. In summary, we make the following contributions:

\begin{itemize}
    \item Extend\footnote{This work is an extension of our conference paper~\cite{GATSBI}. Following are the key differences between the conference version and this version:\begin{enumerate}[labelindent=1cm,leftmargin=*]
        \item Expanded to general infrastructure instead of bridge-only
        \item Implemented crack detection network~\cite{Cheng_2024_CVPR}
        \item Comparison against Structural Inspection Planner~\cite{BABOOMS_ICRA_15}
        \item Experiments on real infrastructure
        \item More robust implementation of viewing cones
        \item Ability to use top or bottom facing cameras with forward-facing
        \item Switched low-level planner to RRT*
        \item Use of depth camera pointclouds in simulation
    \end{enumerate}} our receding horizon algorithm, GATSBI~\cite{GATSBI}, an efficient infrastructure inspection planner with no prior structural information;
    \item Demonstrate that GATSBI outperforms a baseline inspection method that knows the infrastructure surface a priori by \textbf{38\%} through numerous simulations in AirSim using 3D models of infrastructure;
    \item Validate the practical feasibility in experiments with a practical version of GATSBI;
    \item Provide ROS packages that integrate MoveIt with AirSim for simulations and DJI's SDK for real-world experiments.\footnote{\url{https://github.com/raaslab/GATSBI}}
\end{itemize}

\section{Related Work}\label{sec:related}

In this section, we review some of the previous work in this domain. We first describe some of the recent work on the planning side and then describe recent work on crack detection and localization.
\subsection{Planning}

As described earlier, frontier-based exploration is a widely-used method for 3D exploration of unknown environments~\cite{zhu20153d,da2020novel,niroui2017robot}. Other works proposed variants of frontier exploration focused on choosing the next frontier to visit~\cite{dai2020fast,shen2012autonomous}. Another popular approach is to model the exploration problem as one of information gathering and choose a path (or a frontier) that maximizes the information gain~\cite{corah2019communication,premkumar2020combining}. Additionally, there are Next-Best-View approaches~\cite{pito1999solution, prednbv, dhami2024mapnbv} that, as the name suggests, plan the next-best location to take an image from to explore the environment. We refer the reader to a recent, comprehensive survey on multi-robot exploration by Li et al.~\cite{li2020exploration}.
There is also work that uses GTSP for view planning~\cite{tokekar2016algorithms}, however, it focuses on only 2D space and does not take practical issues into consideration such as obstacle avoidance. 

As shown in our conference paper~\cite{GATSBI}, generic exploration strategies are inefficient when performing targeted infrastructure inspection. There has been work on designing inspection algorithms that plan paths that take into account the viewpoint considerations~\cite{peng2019adaptive,hollinger2013active,roberts2017submodular,song2020online}. When prior information is available, one can plan inspection paths carefully by considering the geometric model of the environment. Typically, algorithms use prior information, such as a low-resolution version of the environment, to create an inspection path and obtain high-resolution measurements of the environment~\cite{peng2019adaptive,roberts2017submodular}. Unlike these works, we consider a scenario where the robot has no prior environmental information and must plan using incrementally revealed data.

Bircher et al.~\cite{bircher2018receding} presented a receding horizon planner for exploration and inspection. Both algorithms use a Rapidly-Exploring Random Tree to generate a set of candidate paths in the known, free space of the environment. Then the algorithm selects a path based on a criterion that values how much information a path gains about the environment. The planner uses a receding horizon algorithm repeatedly invoked with new information. We follow a similar approach; however, their algorithm knows the inspection surface a priori, \revone{while our work only requires the UAV to see a portion of the infrastructure at the start of inspection}. The environment is not known for their exploration algorithm but it is known for their inspection algorithm. GATSBI has no prior knowledge about the environment and \revone{minimal knowledge of the} target inspection surface. Furthermore, we cluster potential viewpoints using GTSP which leads to further efficiency. Their inspection planner, Structural Inspection Planner (SIP)~\cite{BABOOMS_ICRA_15}, is the baseline inspection strategy that GATSBI is compared against in Section~\ref{sec:eval}.

Song et al.~\cite{song2020online} recently proposed an online algorithm that consists of a high-level coverage planner and a low-level inspection planner. The low-level planner takes into account the viewpoint constraints and chooses a local path that gains additional information about the structure under inspection. Our work differentiates by guaranteeing the quality of inspection, not requiring a bounding box around the target infrastructure, and segmenting the infrastructure of interest from the environment which guarantees inspection of only the target infrastructure.

\subsection{Crack Detection and Localization}
Detecting defects in the infrastructure with artificial intelligence (AI)  has gained interest recently. Deploying these systems on aerial robots requires consideration of accuracy, inference speed, and lack of data. Early defect detection methods use CNN-based approaches such as YOLO~\cite{alfarrarjeh2018deep, faramarzi2020road, kuang2020computer}, Faster R-CNN~\cite{cha2017deep}, and SSD~\cite{maeda2018road}. more advanced methods have since used more complex architectures based on Graph Neural Networks~\cite{shang2022superpixel} and Transformers~\cite{liu2021crackformer}. Although these approaches can provide good detection and localization accuracy, they also require huge training datasets, which may not be infeasible for specific defects such as cracks in bridges. 
Also, more complex and larger models increase inference time, making them less attractive for real-time applications.

The existing datasets for crack detection such as CrackTree260~\cite{zou2012cracktree},  CrackLS315~\cite{zou2018deepcrack}, Stone331~\cite{konig2021optimized}, etc., contain only a few hundred images. Data augmentation and transfer learning~\cite{nath2021s2d2net} are commonly used to address the lack of data. Most networks for defect inspection predict a binary segmentation map~\cite{zavrtanik2021draem, bovzivc2021end, tabernik2020segmentation, dougan2022new, li2021automatic, liu2021crackformer, inoue2021crack}. Such pixel-to-pixel prediction networks can be computationally intensive and defect localization may need post-processing. Bounding box predictors provide a good alternative~\cite{sun2022new, kuang2020computer, joshi2022automatic}, but the small size of the training data remains a challenge for these networks as well. To address these issues, we use a real-time open-set detector, pre-trained on large-scale data with text-based prompts. Specifically, we use the YOLO-World~\cite{Cheng_2024_CVPR} model with `crack' as the prompt for crack detection and localization. Unlike the previous works, we do not fine-tune this network and demonstrate that it provides appreciable detection and localization over real-world images, without additional training or fine-tuning.

\section{Problem Formulation}\label{sec:prob}

The goal of our unified system is to find infrastructure defects, specifically cracks that can be detected visually, using a UAV. Assuming we have a UAV with a 3D pointcloud sensor and RGB camera, our goal is find a path to inspect every point on the infrastructure surface while minimizing total flight distance.

We consider the scenario where the geometric model of the infrastructure may be unknown a priori. We assume that the UAV starts the algorithm at a location where at least some part of the infrastructure is visible. If this is not the case, we can run a frontier exploration strategy until the infrastructure is visible. We then plan an inspection path for the part of the infrastructure that is visible. As the UAV sees more of the infrastructure, we replan to find a better tour in a receding horizon fashion. 

We use a 3D semantic, occupancy grid built using localized pointcloud data to represent the model of the infrastructure built online. GATSBI assigns each voxel in the occupancy grid a semantic label. The label indicates whether the voxel is free space $v_{F} \in V_{F}$; is occupied space, part of the infrastructure, and previously inspected $v_{BI} \in V_{BI}$; is occupied, part of the infrastructure, but not yet inspected $v_{BN} \in V_{BN}$; and occupied but not a infrastructure voxel (i.e., obstacles)  $v_{O} \in V_{O}$. Our goal is to inspect all the voxels that correspond to the infrastructure surface, i.e., to ensure that $V_{BN} = \emptyset$. 

A voxel $v_{BN} \in V_{BN}$ is inspected if we inspect at least one of its six faces. A face is inspected if the center of that face falls within a cone given apex angle centered at the UAV camera and within a minimum and maximum range of the UAV camera. The apex angle represents the field of view of the camera that is rigidly attached to the UAV. The viewing distance is a minimum and maximum distance range that the UAV should inspect an infrastructure voxel to ensure quality images for inspection. Figure~\ref{fig:vcvd} shows an example of these viewing constraints. 
For the rest of the paper, we refer to the viewing cone and distance as $View$. The RGB camera is used to take pictures of the bridge once the UAV has reached a target $View$ point. The problem we are trying to solve can be defined as follows:

\paragraph*{Problem 1} \label{prob:1}
Given a 3D occupancy map consisting of four sets of voxels ($V_{F}, V_{BI}, V_{BN}, V_{O}$), find a minimum length path that inspects every voxel in $V_{BN}$.

We repeatedly solve Problem 1 as we gain new information until $V_{BN} = \emptyset$. In the Section~\ref{sec:GATSBI}, we show how to model this problem as a GTSP instance.

Given a collection of RGB images captured at each inspection point, we seek to localize defects along the visible surfaces of the infrastructure via two phases offline. The first phase uses a machine learning pipeline to propose potential defects and their locations within each image. In the next phase, a human expert accepts or rejects each proposed defect and determines if the defect requires repairs. We assume the existence, number, and locations of defects are unknown \textit{a priori}.

\section{System Overview} \label{sec:overview}
\begin{figure}
    \centering
    \includegraphics[width = 0.8\columnwidth]{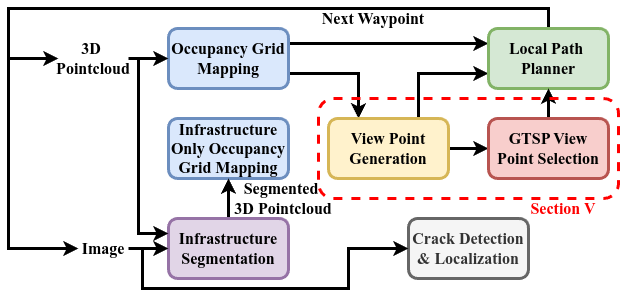}
    \caption{Flow diagram of GATSBI. \revone{The algorithm creates an occupancy map of the environment using incoming 3D pointclouds. Then, it segments the points corresponding to the infrastructure into another point cloud using the RGB camera images. It then makes another occupancy map of only the infrastructure using the segmented point cloud. GATSBI uses both the environment and infrastructure occupancy maps to generate viewpoints, points in free space where the UAV can inspect the infrastructure. It sends these to the GTSP instance to make a tour and then a local path planner to get the flight path.}}
    \label{fig:flowDiagram}
\end{figure}

In this section, we give an overview of the GATSBI algorithm. We show the full pipeline (Fig.~\ref{fig:flowDiagram}) which broadly consists of three perception modules (segmentation, SLAM, and occupancy grid mapping) and two planning modules (high-level GTSP inspection planning and low-level point-to-point planning). Data captured during the flight is fed into the 6th module, crack detection. The main algorithmic contribution, the high-level planner, is described in the next section~\ref{sec:GATSBI}. Here we describe all other off-the-shelf components that form the full system.

\subsection{Perception} \label{sec:perception}

\begin{figure}
    \centering
    \includegraphics[width = 0.75\columnwidth]{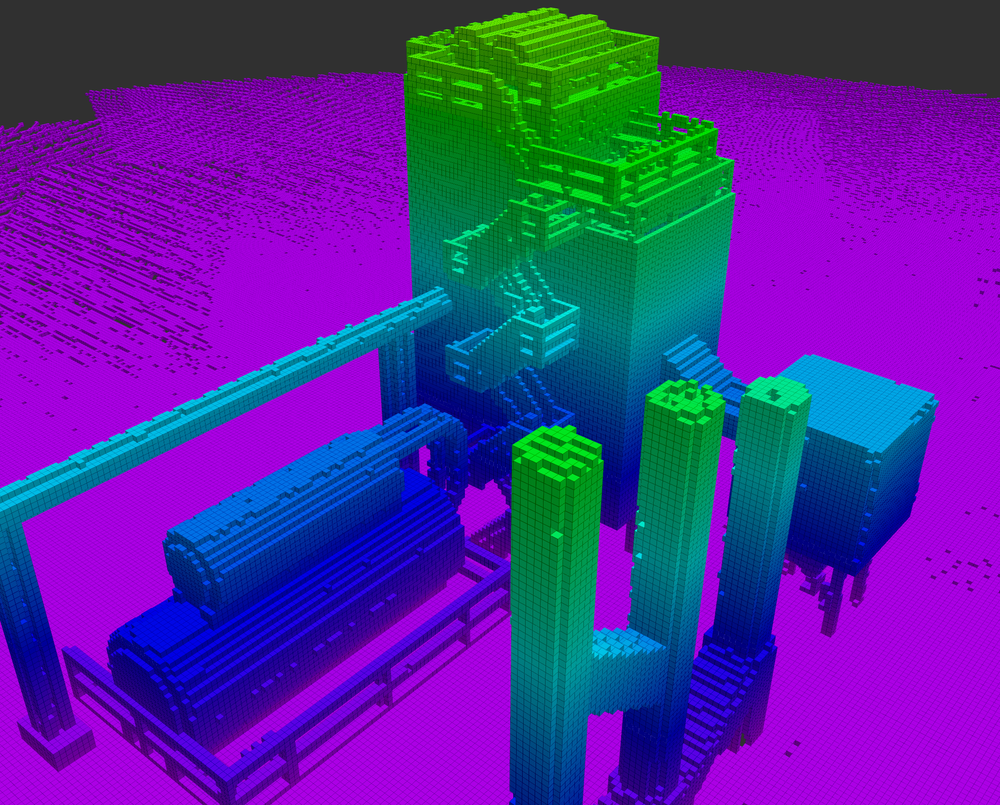}%
    \caption{Full voxel map containing $V_{BI}$ (inspected infrastructure voxels), $V_{BN}$ (uninspected infrastructure voxels), and $V_O$ (obstacle voxels).}%
    \label{fig:fullAndSeg}%
\end{figure}

\subsubsection{Simultaneous Localization and Mapping}
To start, the raw pointcloud data is localized and mapped. Localizing the pointcloud allows for the alignment of each successive pointcloud to all previous ones which is subsequently used to build a map of the environment. As the UAV flies to new areas, more and more of the environment gets mapped including the target infrastructure. 

\subsubsection{Segmentation}
We then use the localized 3D points to segment the infrastructure from the environment. Segmenting out the infrastructure allows the algorithm to differentiate between the infrastructure and obstacles in the environment. This allows only the infrastructure to be inspected as opposed to every object in the environment. The 3D points that lie on the segmented infrastructure are classified as infrastructure points. 

\subsubsection{Occupancy Grid Mapping}
These 3D localized infrastructure points are then used to create a 3D occupancy grid. The map produced by the SLAM module lies in continuous space whereas utilizing occupany grid mapping allows us to discretize the 3D space. In parallel, the complete point cloud (segmented infrastructure and non-segmented obstacle points) is used to generate an environmental 3D occupancy grid. Together, these two occupancy grids output a set of voxels: free $V_F$, bridge $V_{BI}$, bridge $V_{BN}$, and obstacle $V_O$. The algorithm uses the segmented voxels ($V_{BI}, V_{BN}$) to plan inspection paths. The algorithm uses the other voxels ($V_{O}$, $V_{F}$) to plan collision-free paths and take into account viewing constraints. An example of a 3D occupancy grid is shown in Fig.~\ref{fig:fullAndSeg}.

\subsection{Planner} \label{sec:planner}

\subsubsection{High-Level GTSP Inspection Planning}
To inspect infrastructure, we need to inspect all voxels in $V_{BN}$ (as described in Section~\ref{sec:prob}). GATSBI works in a receding horizon fashion. The $V_{BI}$ set keeps track of inspected voxels. This avoids unnecessarily inspecting the same voxel more than once. Specifically, the UAV must view each voxel in $V_{BN}$ from some point on its path within $View$. We formulate this problem as a GTSP instance. The specific details on how we formulate it as a GTSP instance are described in Section~\ref{sec:GATSBI}. 

\subsubsection{Low-Level Point-to-Point Planning}
Once a GTSP tour is received from the high-level planner, point-to-point planning is needed. The GTSP tour does not guarantee collision free paths, it only gives us points where all of the uninspected infrastructure surface can be inspected from. Using our 3D environmental occupancy grid, we can generate collision free paths between each point in the GTSP tour with a point-to-point planner. 

\subsection{Crack Detection}
While the UAV is flying on its inspection path, images from the flight are captured. We can use these images to detect defects in the infrastructure. For our purposes, we focus on crack detection specifically. The raw images can be input into a crack detection neural network where crack locations are output. Since we have a localization and mapping module, the exact location of these cracks can also be tagged in our environment map. 

\section{The GATSBI Planner} \label{sec:GATSBI}
We describe our high-level GTSP inspection planner. As shown in Figure~\ref{fig:flowDiagram}, this is done in two steps: view point generation and GTSP-based view point selection. GTSP generalizes the Traveling Salesperson Problem and is NP-Hard~\cite{Smith2016GLNS}. The input to GTSP consists of a weighted graph, $G$, where vertices are clustered into sets.  The edges of the graph are the distances between the vertices. The objective is a minimum weight tour that visits at least one vertex in each set once. Next, we describe our GTSP setup details. 

\subsection{View Point Generation}
For our implementation, vertices are the center-points of voxels $v_F$ that the UAV can fly to and clusters are the set of all $v_F$ a specific $v_{BN}$ can be inspected from. Each vertex in $G$ corresponds to a candidate viewpoint. We check all pairs of $v_F \in V_F$ and $v_{BN} \in V_{BN}$ to see if $v_F$ lies within $View$ of one of the faces of $v_{BN}$. If so, we add a vertex in the graph $G$ corresponding to the pair $v_F$ and $v_{BN}$. 

\begin{figure}
    \centering
    \includegraphics[width = 0.8\columnwidth]{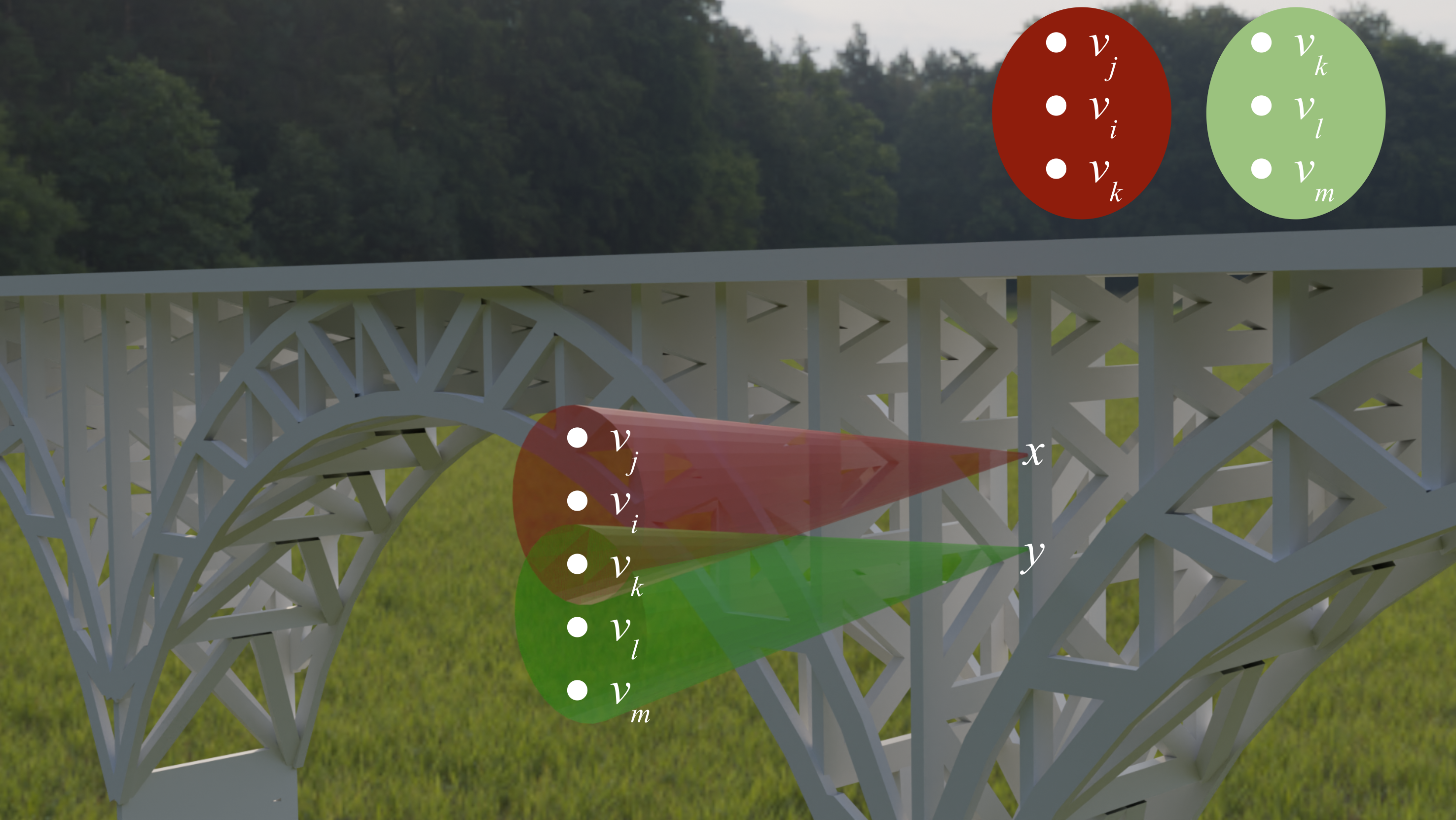}%
    \caption{Example GTSP setup for two infrastructure voxels. Each inspectable infrastructure voxel (x and y) can have multiple potential inspection viewpoints (vertices $v_i$ - $v_m$). All the vertices for a single infrastructure voxel are clustered together (red and green ovals). The edges between these vertices are initially their Euclidean distance.}%
    \label{fig:gtsp_example}%
\end{figure}

Each free voxel that can inspect the same $v_{BN}$ will add one vertex each to the cluster corresponding to $v_{BN}$. A simplified example of this graph setup is shown in Fig.~\ref{fig:gtsp_example}. The GTSP tour will ensure the UAV visits at least one viewpoint in each cluster.

Next, we create an edge between every pair of vertices in $G$. The cost for each of these edges is initially the Euclidean distance between the two vertices.  With the vertices, edges, and clusters, we create a GTSP instance and use the GTSP solver, GLNS~\cite{Smith2016GLNS}, to find a path for the UAV. We use GTSP as our planner because it will guarantee at least one point corresponding to every $V_{BN}$ will be visited.

\subsection{GTSP-Based View Point Selection}
Before moving from one point to the next, we check the distance from the current location of the UAV to the next vertex in the GTSP path. Instead of Euclidean distance, we find the distance of the path between these points using an RRT* algorithm~\cite{rrtstar} with our environment occupancy grid. This ensures the path between these points is collision-free. If the difference between the RRT* distance and the Euclidean distance is greater than $DD$ (discrepancy distance), we update the edge costs in the GTSP instance and replan the GTSP path. We replan as needed to ensure that the first edge in the returned tour is within $DD$ of the Euclidean distance. We call this a lazy evaluation of edge costs. Computing the RRT* distance (which is a more accurate approximation of the actual travel distances) between every pair of vertices would be time-consuming. By checking the discrepancy lazily, we find a tour quickly while also not executing any edge where the actual distance is significantly larger than the expected distance. For our experiments, we set $DD$ to be 125\% of the Euclidean distance to account for some of the variances in paths generated using RRT* algorithms but still allow replanning when necessary.

\begin{algorithm}
\caption{Overview of single iteration of GATSBI}\label{alg:gatsbi}
\begin{algorithmic}[1]
\State{Update occupancy grids with latest localized pointcloud as described in Section~\ref{sec:overview}-\ref{sec:perception}}
\State{Find all inspectable bridge voxels using occupancy grids and remove previously inspected bridge voxels, resulting in $V_{BN}$}
\If{$V_{BN} = \emptyset$}
\State{Terminate}
\EndIf
\State{Create GTSP instance G as described in Section~\ref{sec:GATSBI}}
\While{Difference in the RRT* distance of the first edge in the GTSP solution and its Euclidean distance is greater than a threshold}
\State{Update first edge cost in G with RRT* distance and re-solve GTSP}
\EndWhile
\State{Use RRT* as point-to-point planner for GTSP tour}
\State{Update $V_{BI}$ with latest inspected bridge voxels}
\end{algorithmic}
\end{algorithm}


We track each newly visited cluster during the flight, corresponding to non-inspected infrastructure voxels. The camera captures an image at each visited point for inspection. Once inspected, GATSBI moves these from set $(V_{BN})$ to $V_{BI}$. The plan is executed until either a time limit $RPT$ elapses or the path is completed. We record raw sensor data during the flight and update the inspected and non-inspected voxels afterward. GATSBI terminates when $V_{BN}$ is empty, indicating all infrastructure is inspected. An overview of a single iteration of the GATSBI algorithm is shown in Algorithm~\ref{alg:gatsbi}.

\section{Evaluation}\label{sec:eval}
In this section, we evaluate the algorithm in both simulation and hardware experiments. For the simulations, we compare it against SIP~\cite{BABOOMS_ICRA_15}. Comparison against a baseline frontier-exploration algorithm and parameter tuning was done in our conference paper~\cite{GATSBI}. First, we describe the common setup that was used in both simulations and experiments.

\subsection{Setup}
For both simulations and experiments, we generated the occupancy grid using 3D pointclouds and input them into Octomap. We implemented the RRT* algorithm using the MoveIt software package, which finds collision-free paths using the environmental 3D occupancy grid. We used a viewing cone with a 20° apex angle and a viewing distance of 2-5 meters, ensuring it was within the camera's FoV and safe for UAV flight as suggested by Dorafshan et al. For crack detection, we used YOLO-World to identify cracks in images captured during flight.

\subsection{Simulations}\label{sec:sim}
We present simulation results to evaluate the performance of our proposed unified inspection system. We discuss the setup used to perform simulations and then discuss the inspection environments. Next, we compare GATSBI with SIP~\cite{BABOOMS_ICRA_15} first quantitatively and then qualitatively.

\begin{figure*}
    \begin{minipage}{0.05\linewidth}\centering
        \vspace{1cm}
        \begin{subfigure}[t]{\textwidth}
            \rotatebox[origin=center]{90}{                }
        \end{subfigure}
        \vspace{1cm}
        \begin{subfigure}[t]{\textwidth}
            \rotatebox[origin=center]{90}{GATSBI}
        \end{subfigure}
        \begin{subfigure}[t]{\textwidth}
            \rotatebox[origin=center]{90}{SIP~\cite{BABOOMS_ICRA_15}}
        \end{subfigure}
    \end{minipage}
    \begin{minipage}{0.9\linewidth}\centering
    \begin{subfigure}[t]{0.13\textwidth}
        \begin{subfigure}[t]{\textwidth}
            \includegraphics[width=\textwidth]{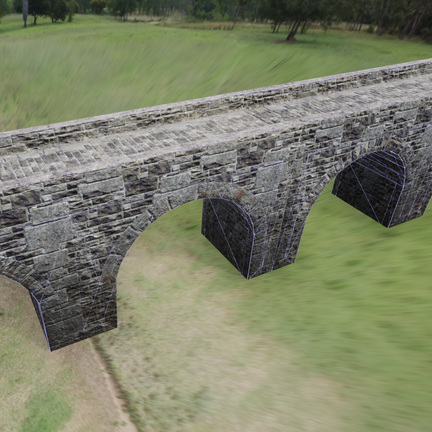}
        \end{subfigure}\vspace{.6ex}
        \begin{subfigure}[t]{\textwidth}
            \includegraphics[width=\textwidth]{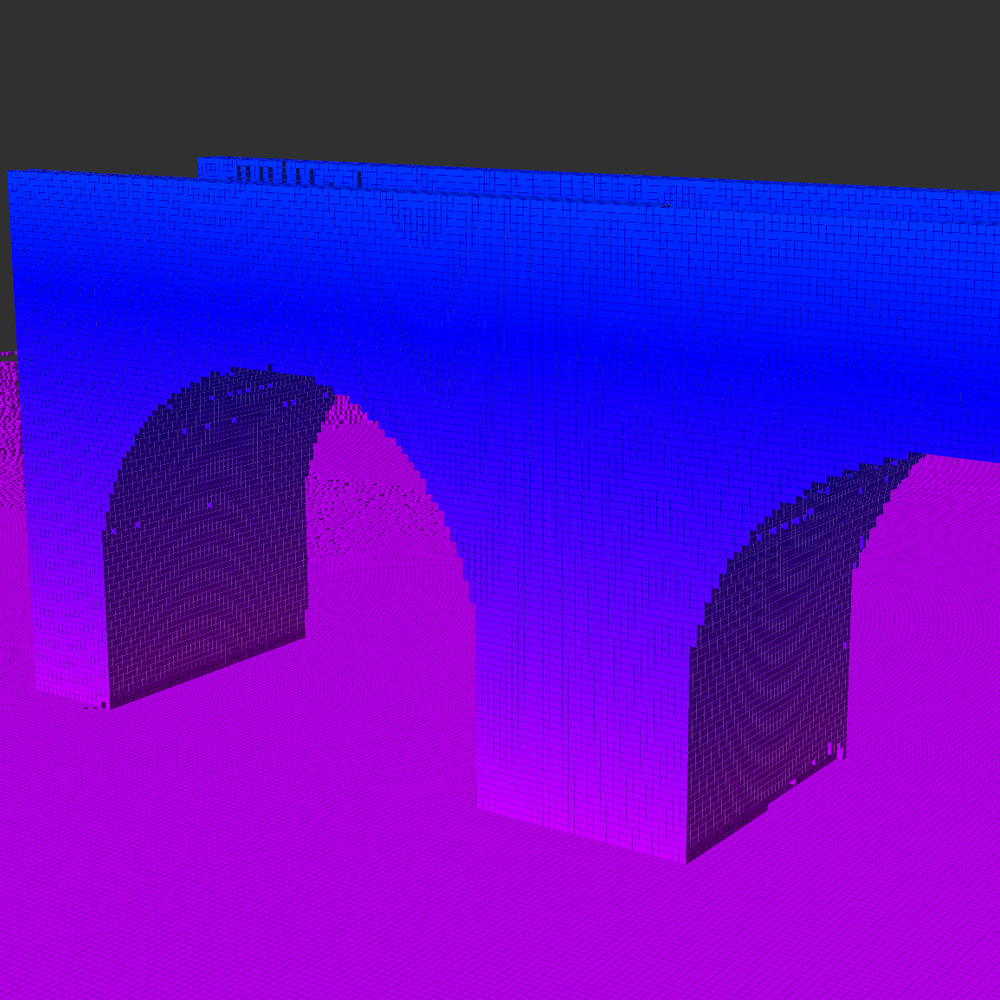}
        \end{subfigure}\vspace{.6ex}
        \begin{subfigure}[t]{\textwidth}
            \includegraphics[width=\textwidth]{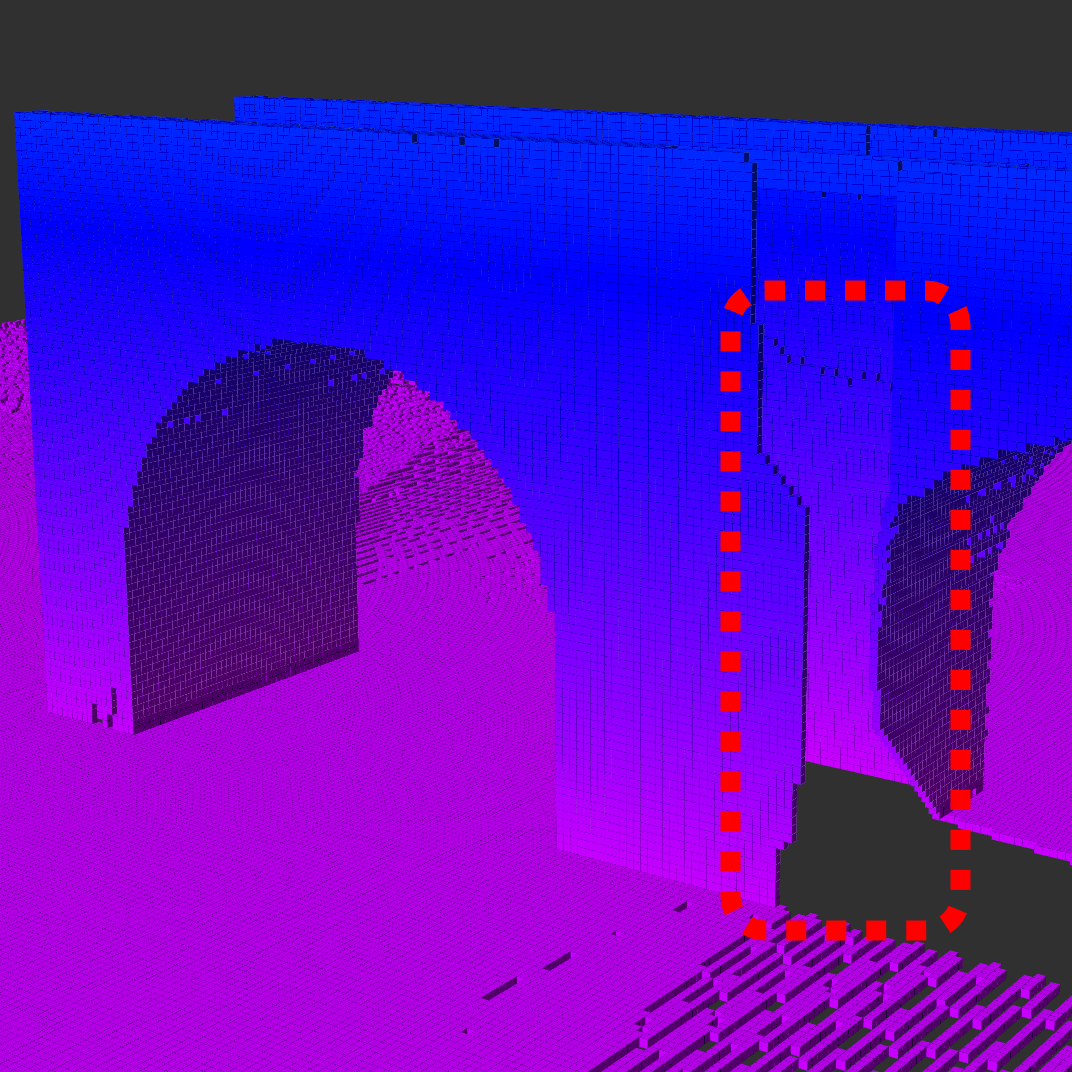}
        \end{subfigure}\vspace{.6ex}
        \caption{Arch}
    \end{subfigure}
    \hfill
    \begin{subfigure}[t]{0.13\textwidth}
        \begin{subfigure}[t]{\textwidth}
            \includegraphics[width=\textwidth]{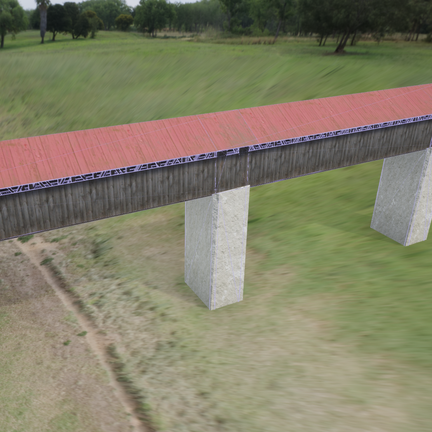}
        \end{subfigure}\vspace{.6ex}
        \begin{subfigure}[t]{\textwidth}
            \includegraphics[width=\textwidth]{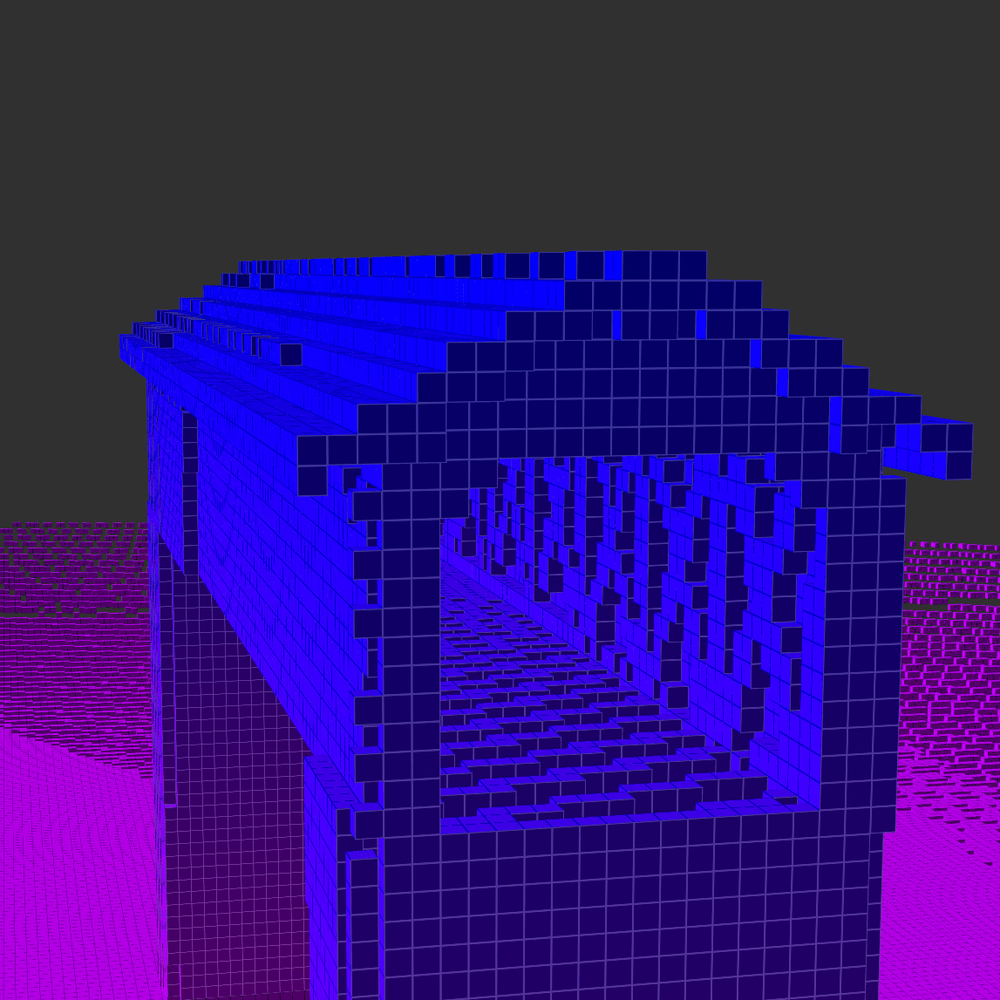}
        \end{subfigure}\vspace{.6ex}
        \begin{subfigure}[t]{\textwidth}
            \includegraphics[width=\textwidth]{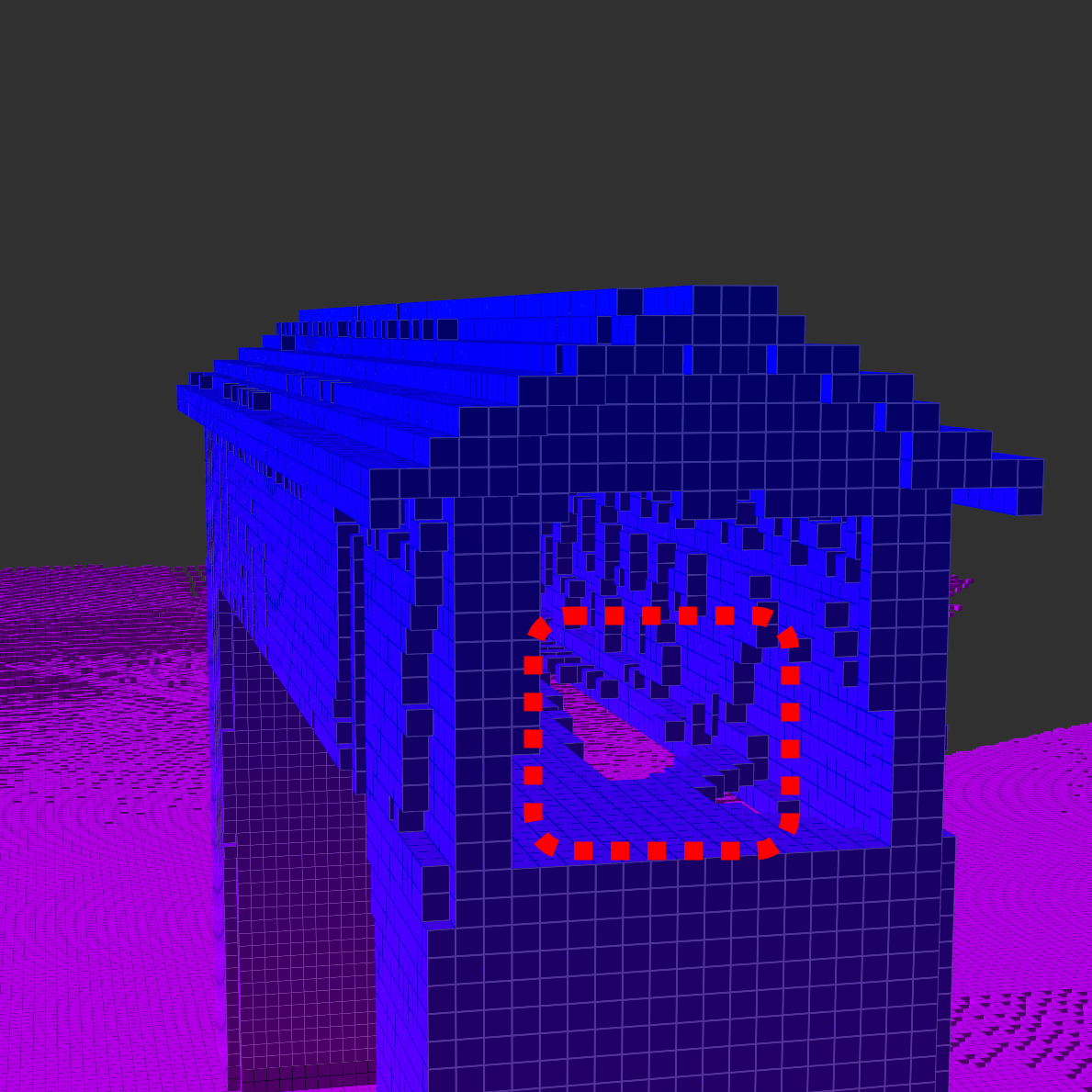}
        \end{subfigure}\vspace{.6ex}
        \caption{Covered}
    \end{subfigure}
    \hfill
    \begin{subfigure}[t]{0.13\textwidth}
        \begin{subfigure}[t]{\textwidth}
            \includegraphics[width=\textwidth]{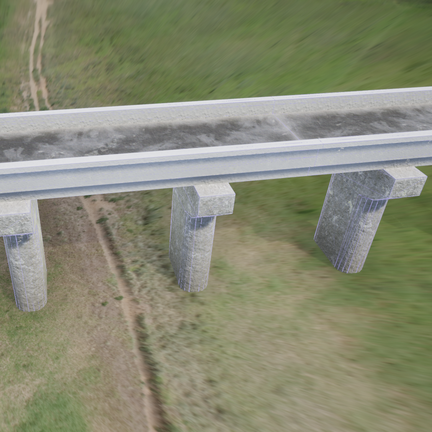}
        \end{subfigure}\vspace{.6ex}
        \begin{subfigure}[t]{\textwidth}
            \includegraphics[width=\textwidth]{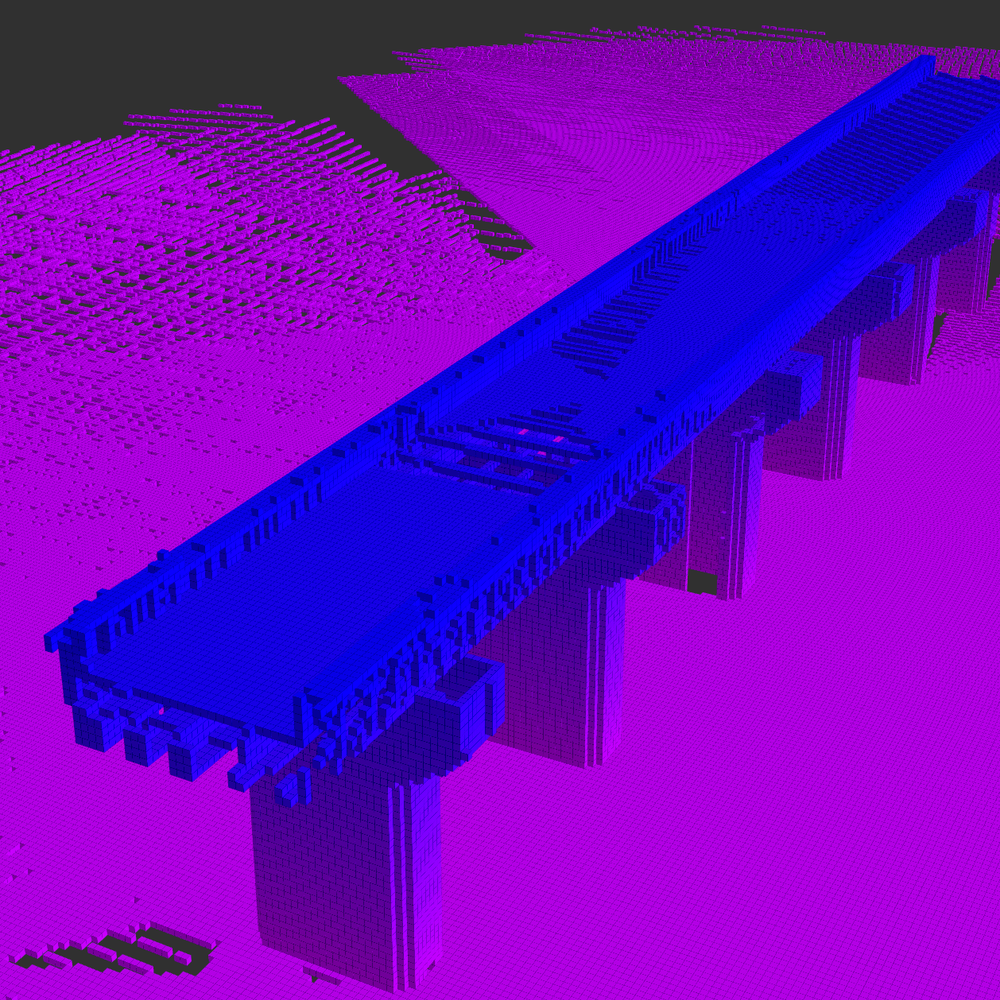}
        \end{subfigure}\vspace{.6ex}
        \begin{subfigure}[t]{\textwidth}
            \includegraphics[width=\textwidth]{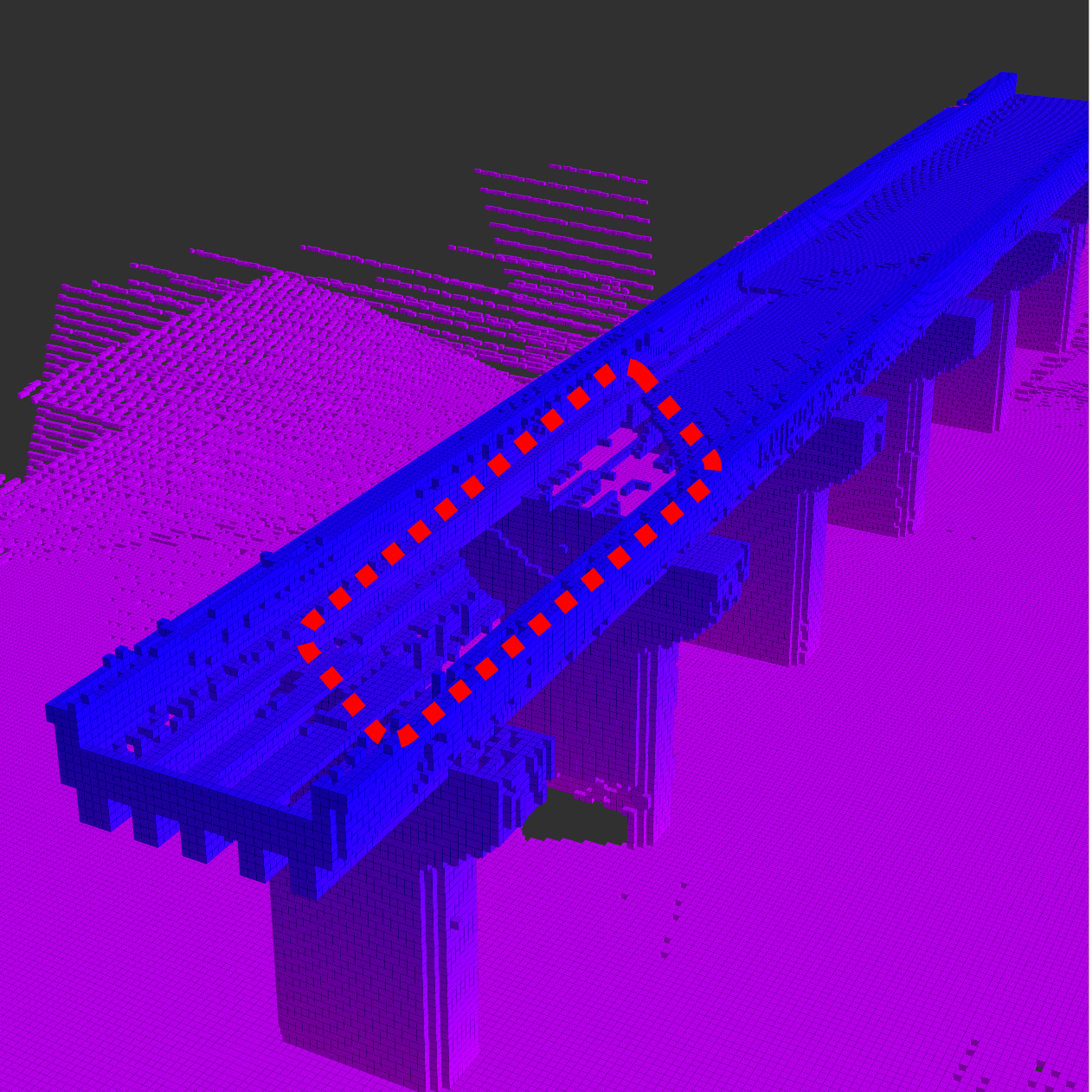}
        \end{subfigure}\vspace{.6ex}
        \caption{Girder}
    \end{subfigure}
    \hfill
    \begin{subfigure}[t]{0.13\textwidth}
        \begin{subfigure}[t]{\textwidth}
            \includegraphics[width=\textwidth]{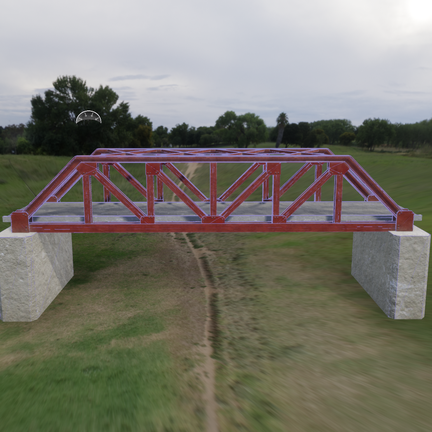}
        \end{subfigure}\vspace{.6ex}
        
        \begin{subfigure}[t]{\textwidth}
            \includegraphics[width=\textwidth]{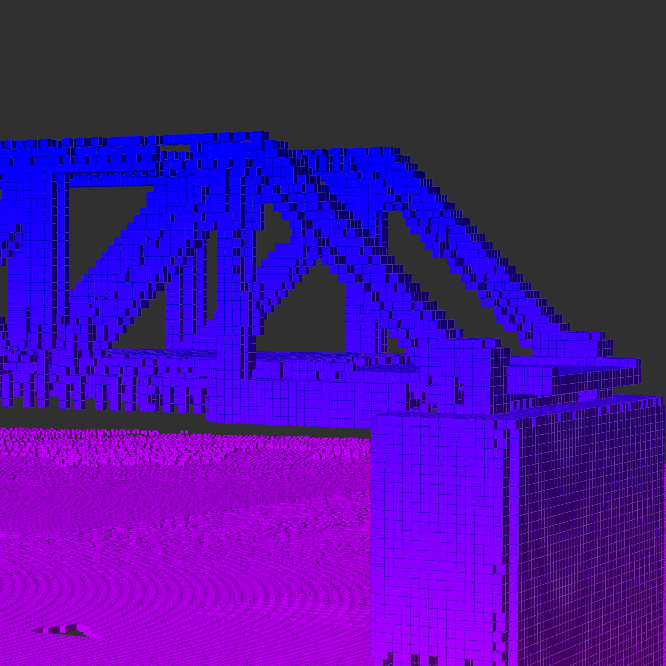}
        \end{subfigure}\vspace{.6ex}
        \begin{subfigure}[t]{\textwidth}
            \includegraphics[width=\textwidth]{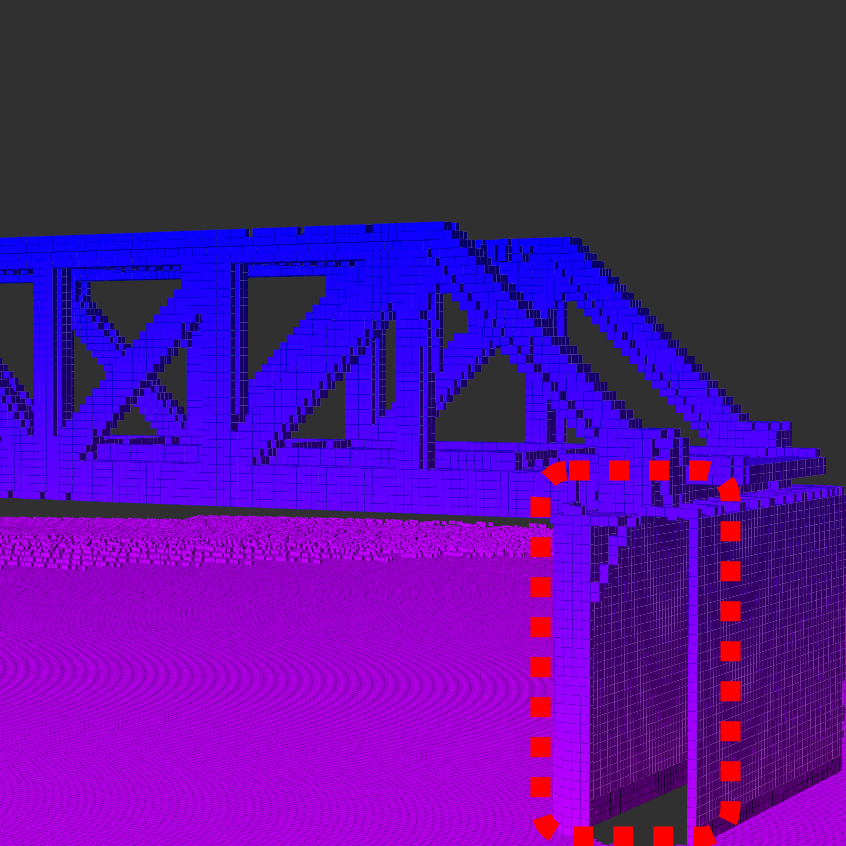}
        \end{subfigure}\vspace{.6ex}
        \caption{Iron}
    \end{subfigure}
    \hfill
    \begin{subfigure}[t]{0.13\textwidth}
        \begin{subfigure}[t]{\textwidth}
            \includegraphics[width=\textwidth]{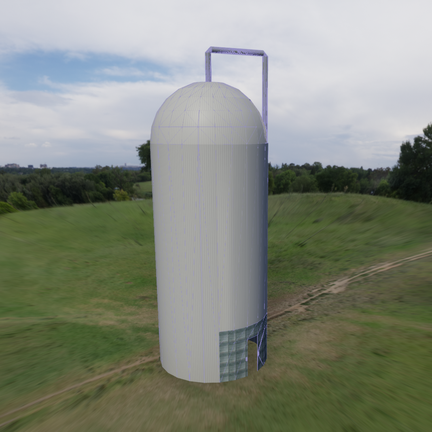}
        \end{subfigure}\vspace{.6ex}
        \begin{subfigure}[t]{\textwidth}
            \includegraphics[width=\textwidth]{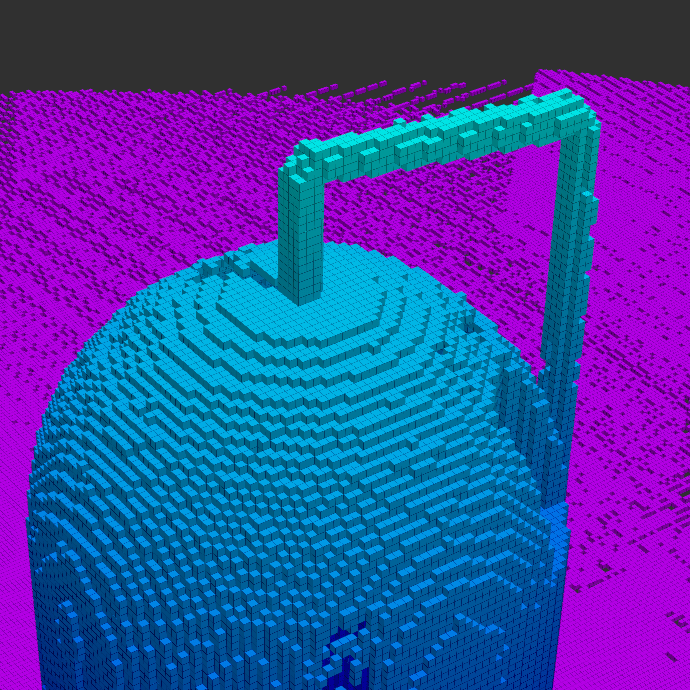}
        \end{subfigure}\vspace{.6ex}
        \begin{subfigure}[t]{\textwidth}
            \includegraphics[width=\textwidth]{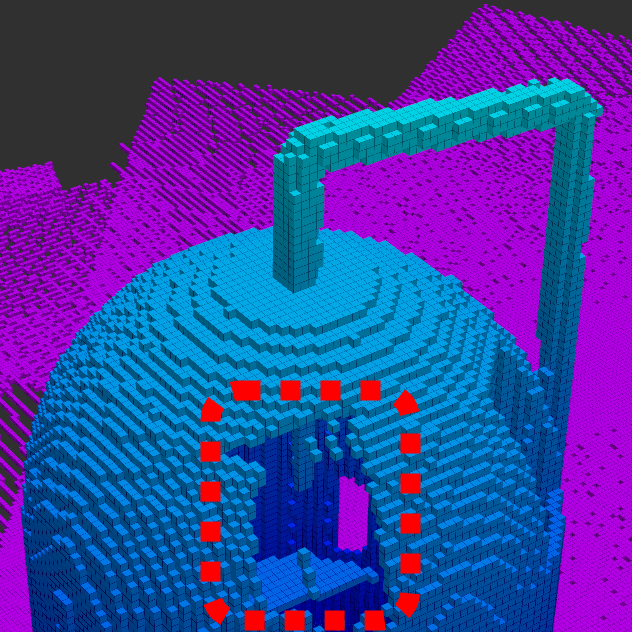}
        \end{subfigure}\vspace{.6ex}
        \caption{Silo}
    \end{subfigure}
    \hfill
    \begin{subfigure}[t]{0.13\textwidth}
        \begin{subfigure}[t]{\textwidth}
            \includegraphics[width=\textwidth]{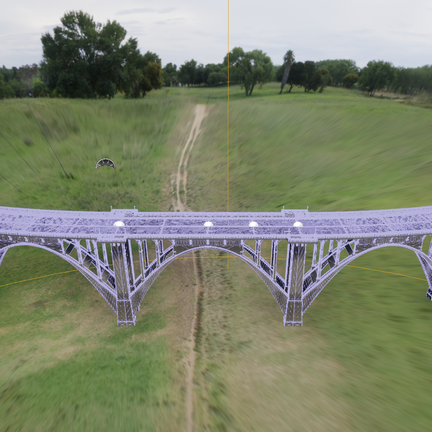}
        \end{subfigure}\vspace{.6ex}
        \begin{subfigure}[t]{\textwidth}
            \includegraphics[width=\textwidth]{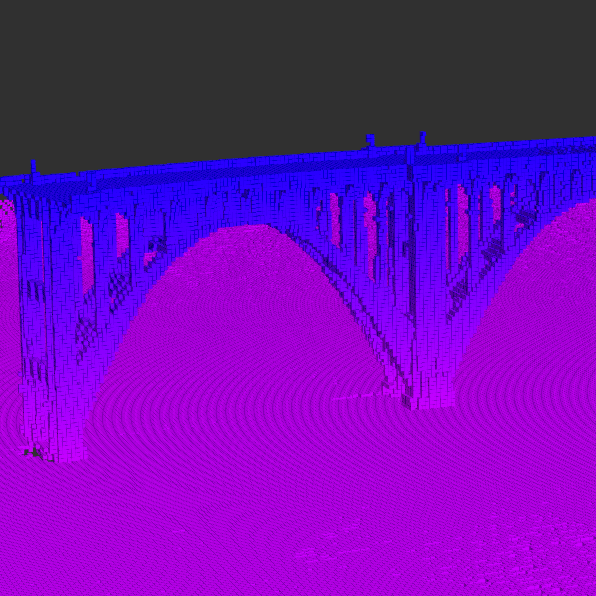}
        \end{subfigure}\vspace{.6ex}
        \begin{subfigure}[t]{\textwidth}
            \includegraphics[width=\textwidth]{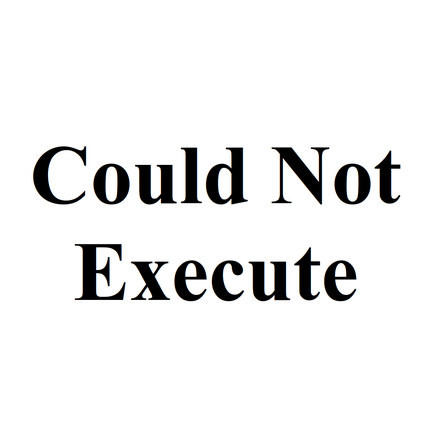}
        \end{subfigure}\vspace{.6ex}
        \caption{Steel}
    \end{subfigure}
    \hfill
    \begin{subfigure}[t]{0.13\textwidth}
        \begin{subfigure}[t]{\textwidth}
            \includegraphics[width=\textwidth]{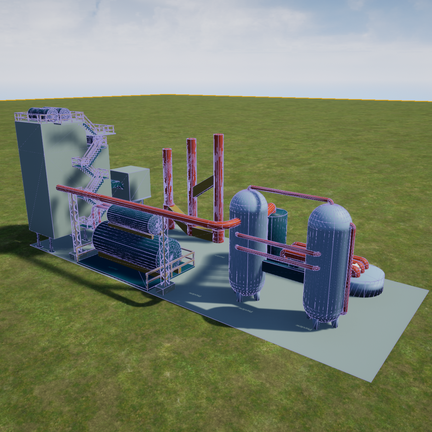}
        \end{subfigure}\vspace{.6ex}
        \begin{subfigure}[t]{\textwidth}
            \includegraphics[width=\textwidth]{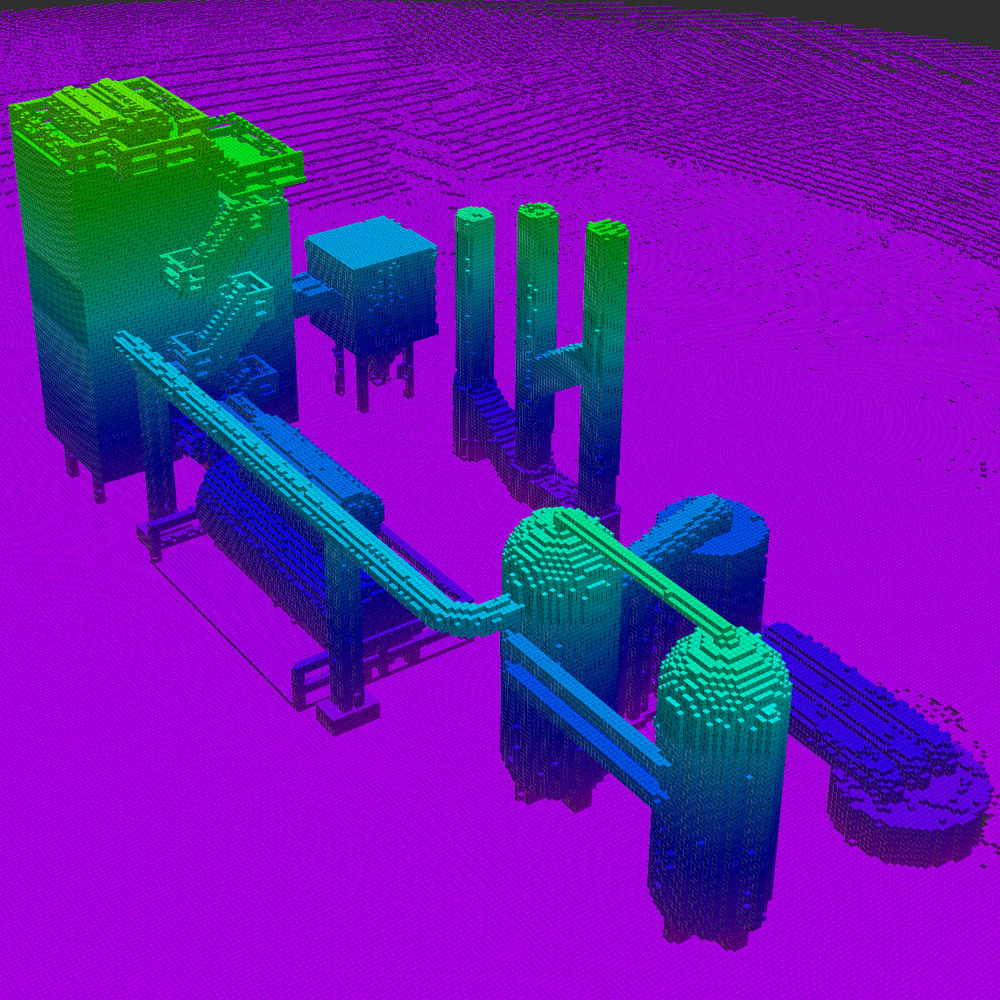}
        \end{subfigure}\vspace{.6ex}
        \begin{subfigure}[t]{\textwidth}
            \includegraphics[width=\textwidth]{figs/Qual_NA.png}
        \end{subfigure}\vspace{.6ex}
        \caption{Plant}
    \end{subfigure}
    \end{minipage}
    \caption{Qualitative comparison between GATSBI (top) and SIP~\cite{BABOOMS_ICRA_15} (bottom) in the seven simulation environments. Top row shows RGB images of the seven different environments. Middle row shows the high-resolution reconstruction after GATSBI's inspection flight. Bottom row shows the high-resolution reconstruction after SIP's reconstruction flight along with coverage gaps shown in the red-dotted boxes.}
    \label{fig:qual_results}
\end{figure*}

\subsubsection{Simulation Setup}
We ran the simulations on a laptop with an Intel Core i9-8950HK CPU, 32 GB of RAM, and an Nvidia RTX 2080 Max-Q GPU running Ubuntu 18.04. All simulated experiments used ROS Melodic and AirSim. Our simulated inspection platform was an AirSim quadrotor equipped with a 512x512 depth camera and an RGB camera. AirSim provided pointcloud localization and built-in semantic segmentation to isolate the infrastructure in the RGB images. The segmented RGB images were aligned with the depth camera to create segmented depth images, which were then converted into 3D pointclouds. This common setup described above was used for the rest of the pipeline.

\subsubsection{Environment Setup} 
\begin{figure}
    \centering
    \includegraphics[width = 1.0\columnwidth]{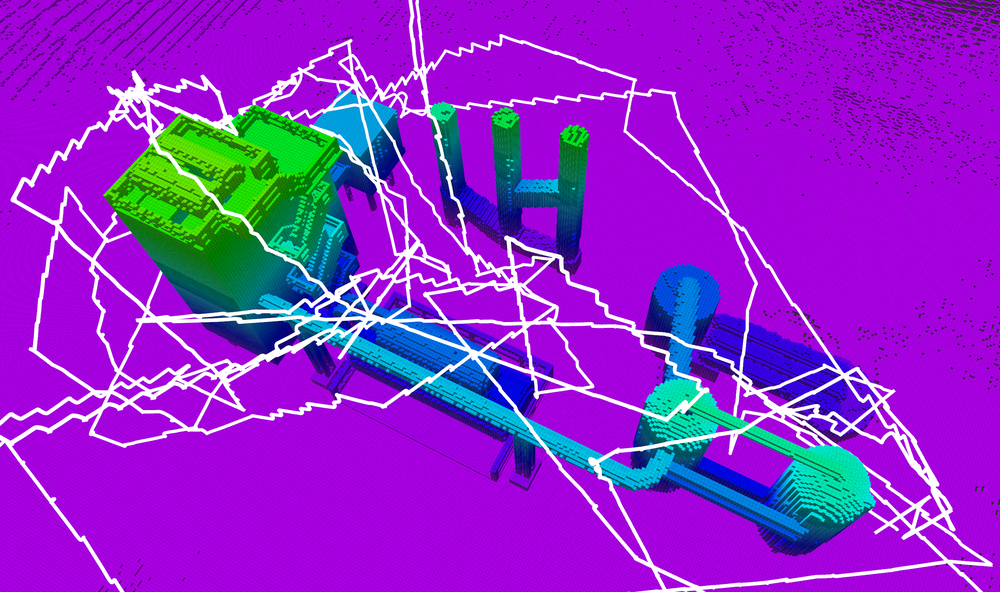}%
    \caption{UAV flight path during GATSBI in the chemical plant environment.}%
    \label{fig:flightPath}%
\end{figure}

We performed experiments using five bridge scenes, a silo scene, and a chemical plant scene (Fig.~\ref{fig:qual_results}). The steel bridge scene contained large hills on both ends of the bridge and trees far away from the bridge. However, the other scenes only contained the infrastructure itself. We chose these infrastructures (bridges and buildings) because they represent distinct types of infrastructure that are common to inspect. Figure~\ref{fig:flightPath} shows a path GATSBI generated and the UAV followed around the chemical plant scene.

\subsubsection{Comparison with SIP}
We compared the inspection performance of GATSBI with SIP. SIP uses a triangular mesh of the target infrastructure and viewing parameters (camera FoV and minimum/maximum viewing distance) to generate viewpoints and paths for inspection. We tested SIP in AirSim, tracking the number of inspected infrastructure voxels during the flight. GATSBI, in contrast, generates a path dynamically as it observes more of the infrastructure, without prior knowledge of the infrastructure map. Both algorithms were evaluated in five scenes: arch bridge, covered bridge, box girder bridge, iron bridge, and silo tower. On average, GATSBI inspected 38\% more voxels than SIP, but SIP generally completed inspections faster and with shorter flight distances due to its prior knowledge of the infrastructure mesh. The results are shown in Table \ref{tab:sim_results}. 

\begin{table}[ht!]
    \centering
    \begin{tabular}{|c|c|c|c|c|c|}
        \hline
        \multirow{2}{*}{\textbf{Scene}} & \multirow{2}{*}{\textbf{Algorithm}} & \textbf{Object} & \multirow{2}{*}{\textbf{Inspected}} & \textbf{Time} & \textbf{Distance} \\
        & & \textbf{Voxels} & & (min) & (meters) \\
        \hline
        \multirow{2}{*}{Arch} & SIP & \multirow{2}{*}{610} & 266 & \textbf{42.29} & \textbf{167.33} \\
        \cline{2-2}\cline{4-6}
         & GATSBI & & \textbf{610} & 91.32 & 295.88 \\
         \hline
        \multirow{2}{*}{Covered} & SIP & \multirow{2}{*}{79} & 70 & 32.22 & \textbf{89.07}  \\
        \cline{2-2}\cline{4-6}
        & GATSBI & & \textbf{79} & \textbf{21.36} & 89.12 \\
        \hline
        \multirow{2}{*}{Girder} & SIP & \multirow{2}{*}{529} & 250 & \textbf{61.02} & \textbf{267.87} \\
        \cline{2-2}\cline{4-6}
         & GATSBI & & \textbf{529} & 84.52 & 339.99 \\
        \hline
        \multirow{2}{*}{Iron} & SIP & \multirow{2}{*}{331} & 234 & 135.03 & 376.76 \\
        \cline{2-2}\cline{4-6}
         & GATSBI & & \textbf{331} & \textbf{58.90} & \textbf{225.63} \\
        \hline
        \multirow{2}{*}{Silo} & SIP & \multirow{2}{*}{481} & 288 & \textbf{83.41} & \textbf{206.53} \\
        \cline{2-2}\cline{4-6}
        & GATSBI & & \textbf{481} & 84.71 & 293.38 \\
        \hline
    \end{tabular}
    \caption{Table showing the number of inspectable voxels that were inspected by GATSBI and SIP for five scenes as well as the total algorithm runtime and flight distance.}
    \label{tab:sim_results}
\end{table}

For a direct comparison, we measured the percentage of voxels inspected by GATSBI at the time and distance when SIP finished. For instance, if SIP finished in 30 minutes or flew 200 meters, we compared the number of voxels GATSBI inspected at 30 minutes or 200 meters. These results are shown in Table \ref{tab:sim_results_equal}. When equalized for time, GATSBI inspected 34\% more voxels on average than SIP. When equalized for distance, GATSBI inspected 35

\begin{table}[ht!]
    \centering
    \begin{tabular}{|c|c|c|c|}
        \hline
        \multirow{2}{*}{\textbf{Scene}} & \multirow{2}{*}{\textbf{Algorithm}} & \textbf{\% Inspected} & \textbf{\% Inspected} \\
        & & \textbf{Equal Time} & \textbf{Equal Distance}\\
        \hline
        \multirow{2}{*}{Arch} & SIP & 43.61\% & 43.61\% \\
        \cline{2-4}
         & GATSBI & \textbf{54.56\%} & \textbf{75.14\%}  \\
         \hline
        \multirow{2}{*}{Covered} & SIP & 82.28\% & 88.61\% \\
        \cline{2-4}
         & GATSBI & \textbf{100.00\%} & \textbf{99.98\%} \\
        \hline
        \multirow{2}{*}{Girter} & SIP & 47.26\% & 47.26\%  \\
        \cline{2-4}
         & GATSBI & \textbf{88.48\%} & \textbf{91.12\%} \\
        \hline
        \multirow{2}{*}{Iron} & SIP & 36.25\% & 42.90\% \\
        \cline{2-4}
         & GATSBI & \textbf{100.00\%} & \textbf{100.00\%} \\
        \hline
        \multirow{2}{*}{Silo} & SIP & 59.88\% & 59.88\% \\
        \cline{2-4}
        & GATSBI & \textbf{100.00\%} & \textbf{92.96\%} \\
        \hline
    \end{tabular}
    \caption{Table comparing the percentage of inspectable voxels that were inspected by GATSBI and SIP for five scenes when equalizing for time and distance.}
    \label{tab:sim_results_equal}
    \vspace{-3mm}
\end{table}

Intuitively, SIP might seem better since it has prior access to the infrastructure's mesh, but GATSBI performs better for several reasons. First, using voxels allows us to define the inspection resolution more easily. A mesh being inspected does not guarantee all its voxels are inspected. Second, SIP does not consider collision-free navigation and may generate inefficient paths since it only knows the infrastructure's mesh, not other obstacles. Lastly, SIP does not check if the mesh can be navigated, potentially leading to unreachable inspection points. GATSBI, however, generates collision-free paths and selects viewpoints based on the entire environment map.

Figure~\ref{fig:qual_results} shows qualitative comparisons between the two methods. GATSBI observes the entire inspection surface, while SIP has coverage gaps (dotted-red boxes). GATSBI also inspects more complex structures that SIP fails to handle due to the complexity of the triangular mesh.
    
\subsection{Real-World Experiments}

This section presents real-world experiments that evaluate the performance of GATSBI. Here, we show results with only GATSBI since our simulated experiments demonstrated that GATSBI outperformed SIP. Below, we present quantitative results that compare the computation time, flight time, and the number of voxels inspected with GATBSI. 

\subsubsection{Experiment Setup} 
We conducted these experiments using a DJI Matrice M600 Pro (see Fig.~\ref{fig:flight}), equipped with an NVIDIA Jetson TX2, Velodyne VLP-16 3D LiDAR, and GPS. Due to energy constraints, we used a practical version of GATSBI, as discussed in Section~\ref{sec:conclusion}. The UAV's LiDAR created a bridge map, and GATSBI ran offline to determine the inspection path. Unlike simulations using depth camera pointclouds, we used 3D LiDAR pointclouds. Initial localization was achieved with DJI's GPS-based system, and for real-world noise, we used LIO-SAM~\cite{liosam2020shan} with GPS. We manually segmented the target infrastructure geographically, which sufficed for our experiments. The common setup was used for the rest of the pipeline.


\subsubsection{UMD F3}

\begin{figure}[ht!]
\vspace{2.5mm}
    \centering
    \begin{subfigure}[b]{0.49\columnwidth}%
        \includegraphics[height = 3.1cm]{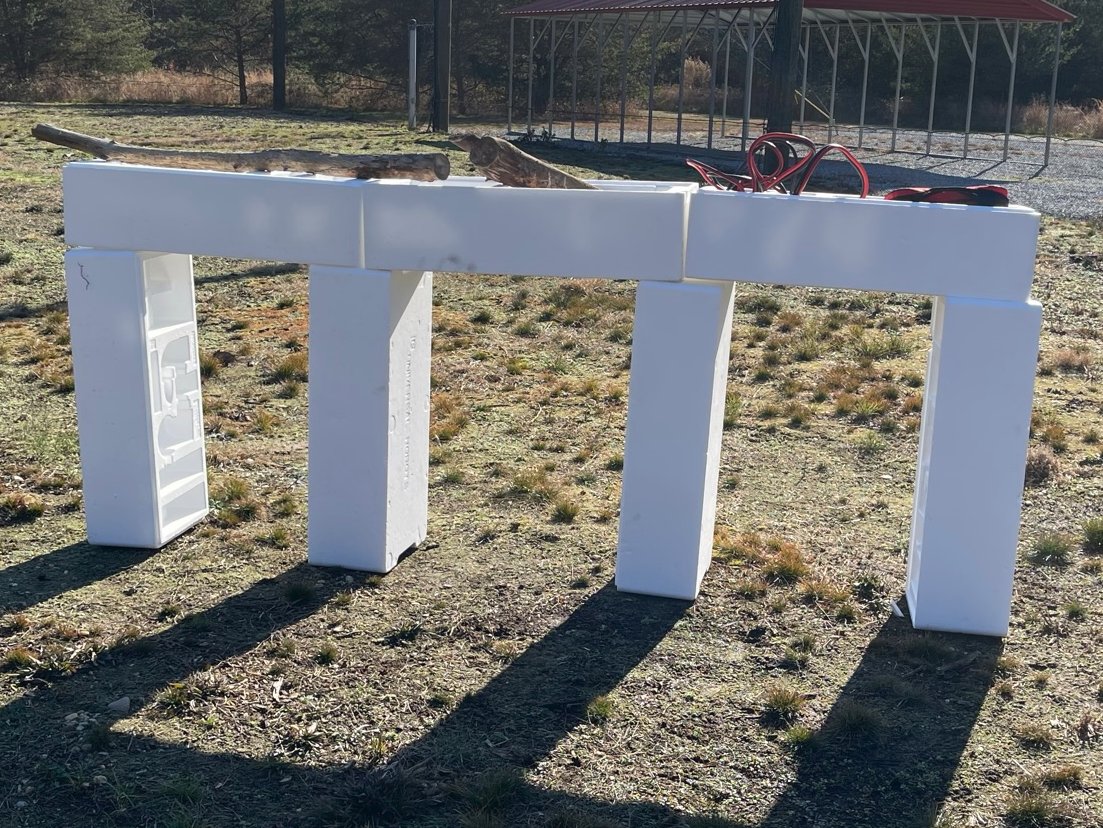}%
    \end{subfigure}%
    \hfill%
    \begin{subfigure}[b]{0.49\columnwidth}%
        \includegraphics[height = 3.1cm]{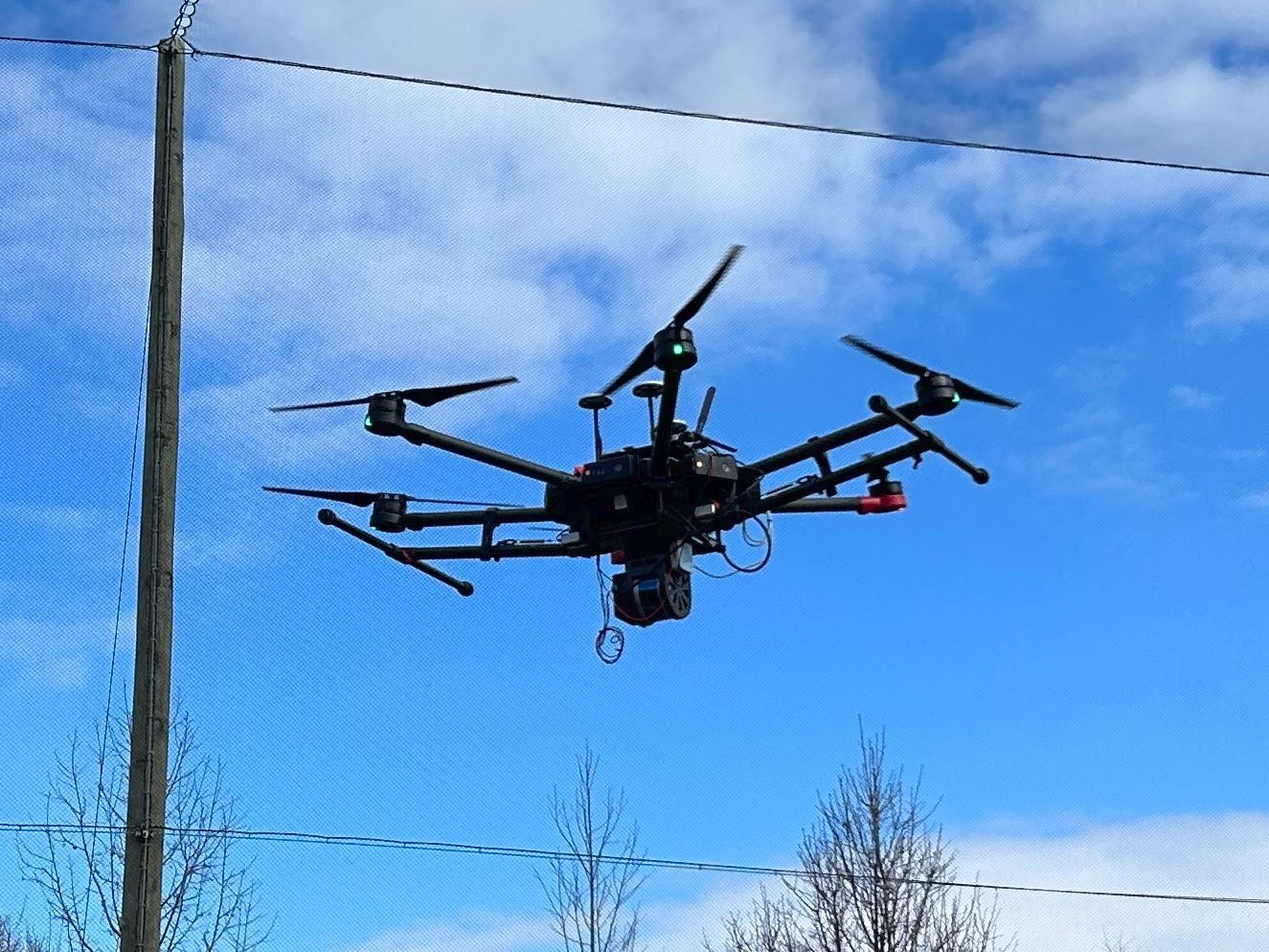}%
    \end{subfigure}%
    \caption{Left: The mock bridge used for the UMD F3 experiments. Right: DJI M600 Pro mid-flight during inspection.}
    \label{fig:flight}
\end{figure}

\begin{figure}[ht!]
    \centering
    \includegraphics[width = 0.85\columnwidth]{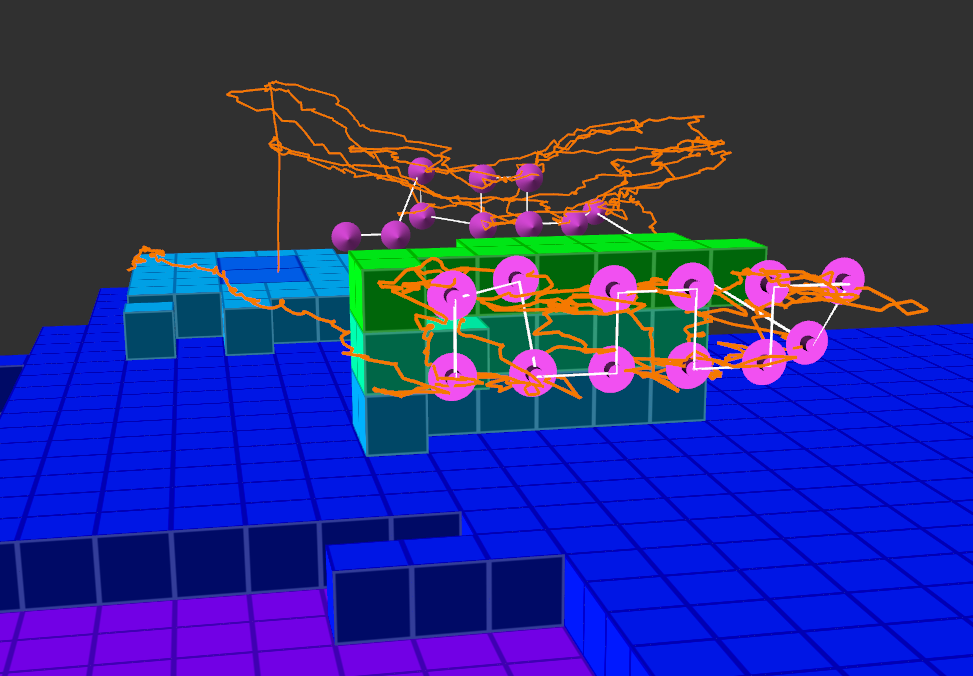}%
    \caption{Flight path and target inspection path on the real-world mock bridge.}%
    \label{fig:paths_hardware}%
\end{figure}

We tested GATSBI on a mock bridge at the Fearless Flight Facility (F3) at the University of Maryland, College Park (see Fig.~\ref{fig:flight}). The UAV flew around the bridge with a viewing cone angle of 0° and a distance range of 2-5 meters, inspecting all 15 bridge voxels. The target inspection path and actual flown path are shown in Fig.~\ref{fig:paths_hardware}, with the white line indicating the direct path between inspection points and the orange line showing the actual flown path. Pink cones represent inspection points. This experiment demonstrates GATSBI's effectiveness for real-world infrastructure inspection. The GATSBI algorithm's computation time was 0.32 seconds, GTSP time was 8.25 seconds, and flight time was 772 seconds, confirming that the algorithm time is not a bottleneck.

\subsubsection{College Park Bridge}

\begin{figure}[ht!]
    \centering
    \begin{subfigure}[b]{0.49\columnwidth}%
        \includegraphics[height = 3.15cm]{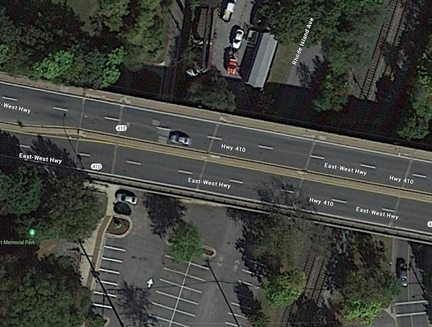}%
    \end{subfigure}%
    \hfill%
    \begin{subfigure}[b]{0.49\columnwidth}%
        \includegraphics[height = 3.15cm]{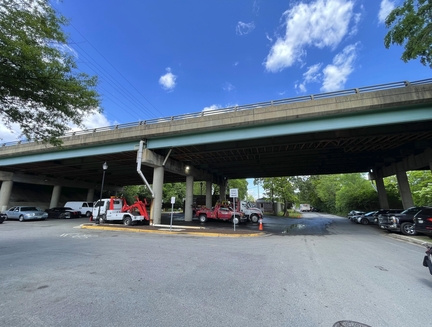}%
    \end{subfigure}%
    \caption{Left: Google Earth image of College Park Bridge. Right: Ground View of College Park Bridge.}
    \label{fig:college_park}
\end{figure}

\begin{figure}
    \centering
    \begin{subfigure}[t]{0.49\columnwidth}
        \includegraphics[width=\textwidth]{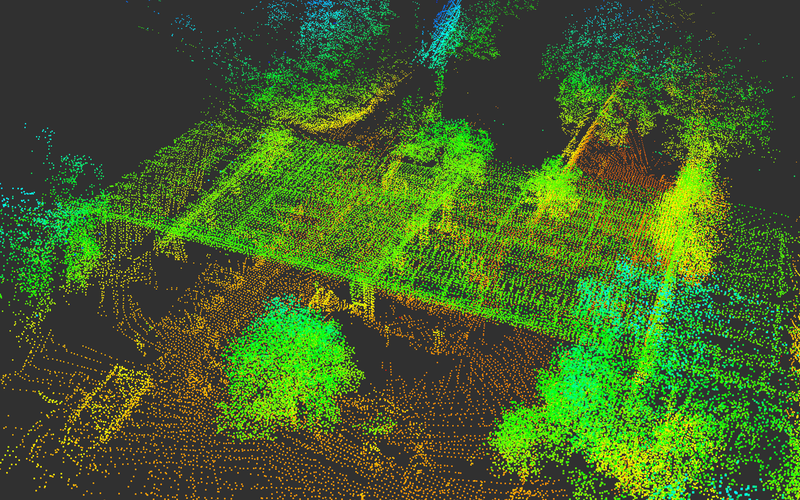}
        \caption{SLAM Pointcloud}
    \end{subfigure}
    \begin{subfigure}[t]{0.49\columnwidth}
        \includegraphics[width=\textwidth]{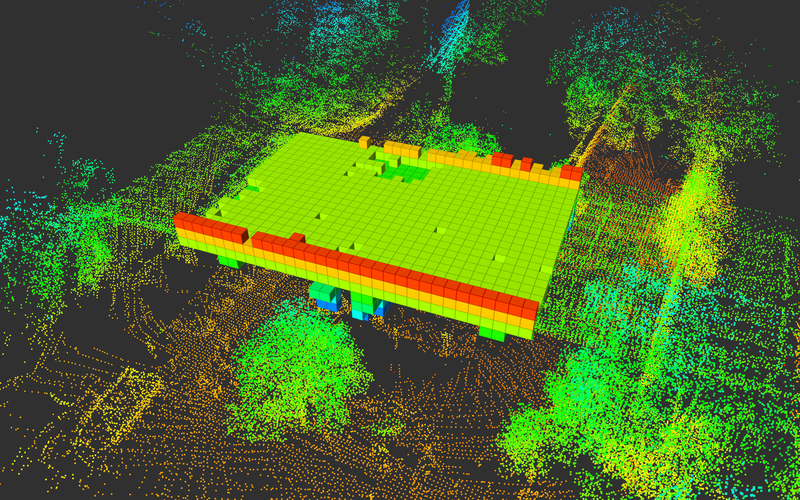}
        \caption{Segmented bridge Octomap}
    \end{subfigure}\vspace{.6ex}
    \caption{SLAM pointcloud and segmented bridge Octomap of real-world College Park Bridge.}
    \label{fig:college_park_points}
    \vspace{-2mm}
\end{figure}

\begin{figure}
    \centering
    \begin{subfigure}[t]{0.85\columnwidth}
        \includegraphics[width=\textwidth]{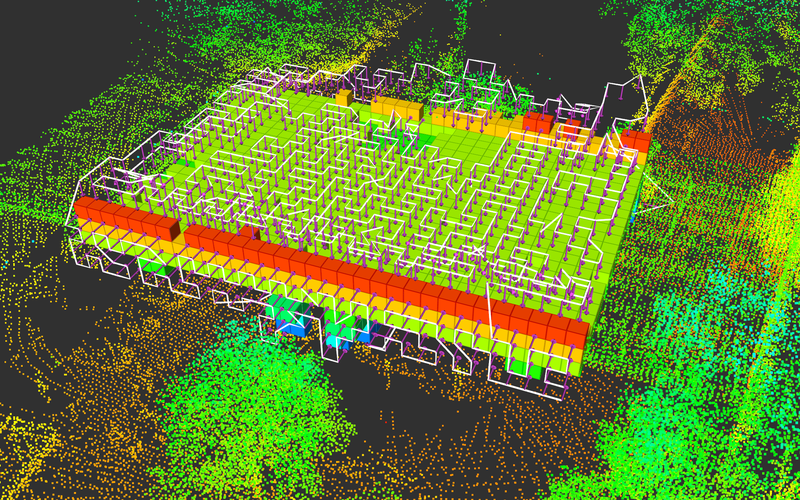}
        \caption{Inspection Path with Viewing Cone 0\degree and 2-5 meter distance}
    \end{subfigure}
    \begin{subfigure}[t]{0.85\columnwidth}
        \includegraphics[width=\textwidth]{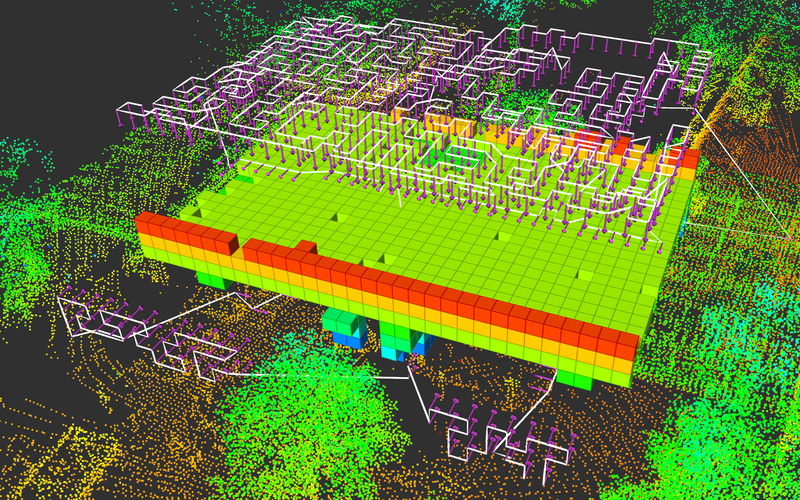}
        \caption{Inspection Path with Viewing Cone 0\degree and 8-10 meter distance}
    \end{subfigure}
    \begin{subfigure}[t]{0.85\columnwidth}
        \includegraphics[width=\textwidth]{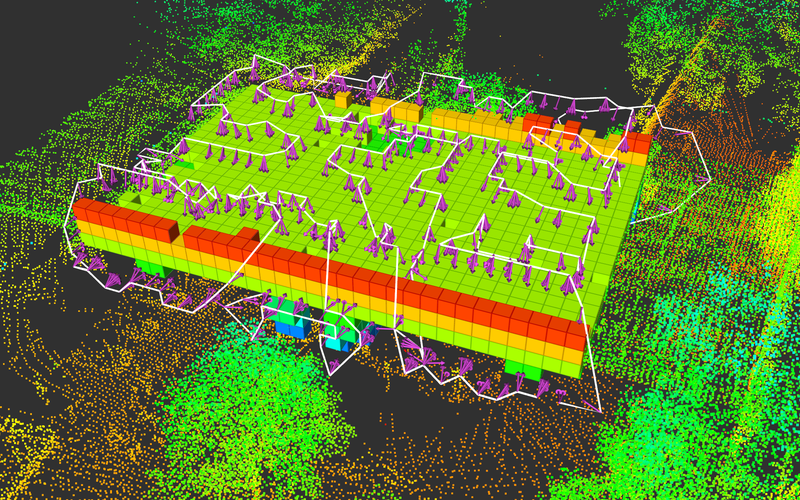}
        \caption{Inspection Path with Viewing Cone 20\degree and 2-5 meter distance}
    \end{subfigure}
    \caption{Output from practical GATSBI algorithm on real-world College Park bridge in 3 different viewing cone scenarios.}
    \label{fig:college_park_gatsbi}
    \vspace{-5mm}
\end{figure}

We tested our method on real infrastructure using the same hardware setup. Due to Washington DC's no-fly zone, we collected data on the ground by placing the DJI M600 Pro in a cart and walking it around a bridge in College Park, Maryland (see Fig.~\ref{fig:college_park}). The data was processed through our pipeline, producing a SLAM pointcloud and a segmented bridge octomap (see Fig.~\ref{fig:college_park_points}). These were used by our practical GATSBI algorithm to generate inspection paths. We modified GATSBI to inspect each voxel face separately. Figure~\ref{fig:college_park_gatsbi} shows three example paths: the first with a 0° viewing cone angle at 2-5 meters, the second with the same angle at 8-10 meters (avoiding a tree), and the third with a 20° angle at 2-5 meters. Multiple arrows at each inspection point indicate multiple faces inspected due to the viewing cone angle.

%


\subsection{Crack Detection Results}
\begin{figure}
    \centering
    \begin{subfigure}[t]{0.49\columnwidth}
        \includegraphics[width=\textwidth]{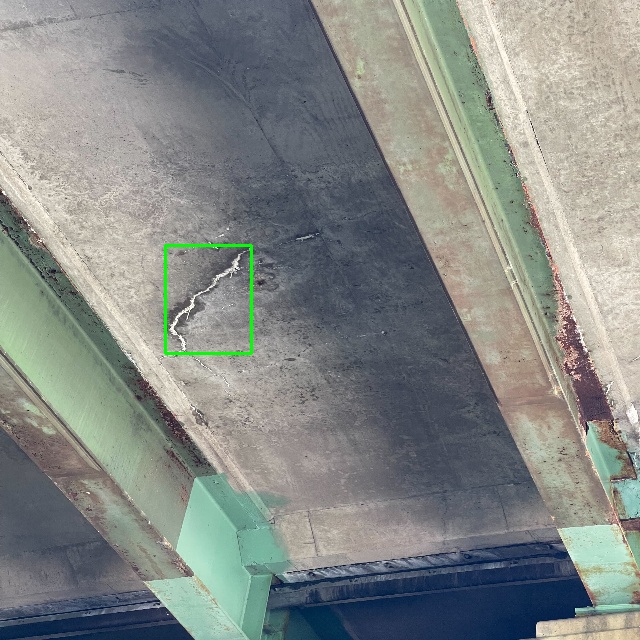}
        \caption{}
    \end{subfigure}
    \begin{subfigure}[t]{0.49\columnwidth}
        \includegraphics[width=\textwidth]{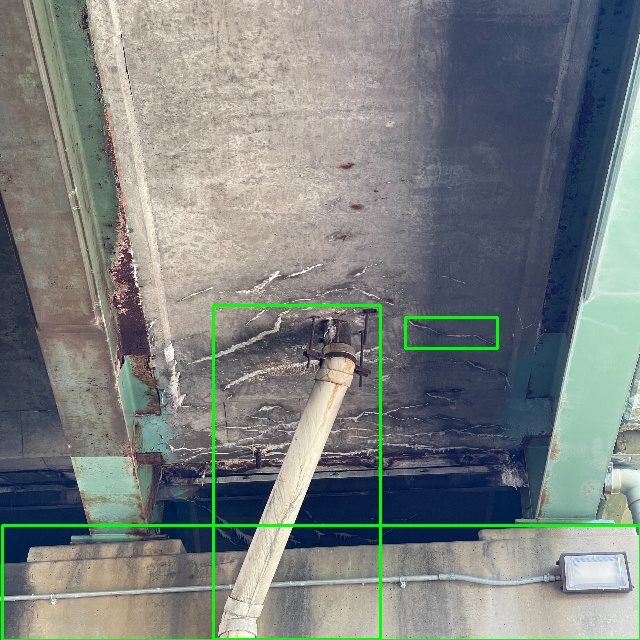}
        \caption{}
    \end{subfigure}
    \begin{subfigure}[t]{0.49\columnwidth}
        \includegraphics[width=\textwidth]{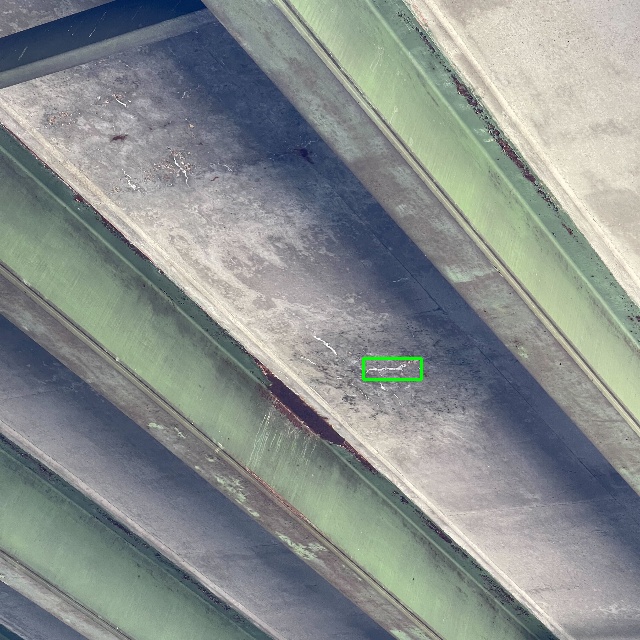}
        \caption{}
    \end{subfigure}
    \begin{subfigure}[t]{0.49\columnwidth}
        \includegraphics[width=\textwidth]{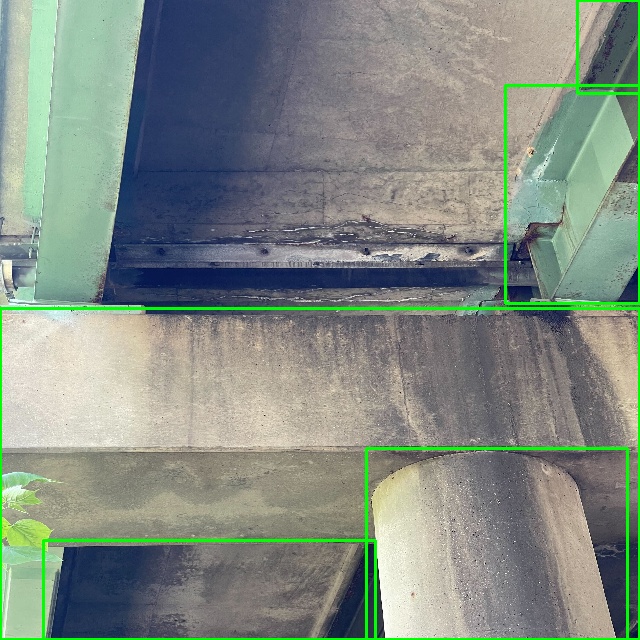}
        \caption{}
        \label{fig:cracks:pillar1}
    \end{subfigure}
    \caption{Crack detection results over real-world College Park bridge images.}
    \label{fig:college_park_cracks}
\end{figure}

For crack detection, we use pre-trained YOLO-World~\cite{Cheng_2024_CVPR} large model (\textit{YOLOw-l}) conditioned on the text prompt `crack'. This crack detection model effectively acts in a zero-shot manner for this task. We run this model on some images from the College Park Bridge, as shown in Figure~\ref{fig:college_park_cracks}. Here we show results with a confidence threshold of 0.008.

We found that YOLO-World can appreciably detect and localize the cracks in the bridge. In some cases, some false positives may occur which may have a high likelihood of containing cracks, such as the pillars in Figure~\ref{fig:cracks:pillar1}. Note that these results were obtained using a pre-trained model and were not fine-tuned for crack detection, highlighting its generalizability. 


\subsection{Code Release}
We provide the GATSBI code as well as two ROS packages that integrate MoveIt in our GitHub repository\footnote{\url{https://github.com/raaslab/GATSBI}}. The GATSBI code is used to run our inspection planner on target infrastructure. One of the ROS packages integrates MoveIt with AirSim's ROS wrapper allowing the use of MoveIt in the AirSim simulation environment. The second ROS package integrates MoveIt with DJI's SDK allowing the use of MoveIt with DJI multirotors in real environments.

\section{Conclusion}\label{sec:conclusion}
We present GATSBI, a 3D infrastructure inspection planner. We evaluate the performance of the algorithm through AirSim simulations and real-world hardware experiments with a UAV equipped with a 3D LiDAR and an RGB camera. The simulations show that GATSBI outperforms SIP~\cite{BABOOMS_ICRA_15}. The hardware experiments show that GATSBI is a viable solution to real-world infrastructure inspection. In particular, we show that the algorithm is efficient in the sense that it targets inspectable voxels rather than simply exploring a volume. The simulations and experiments also demonstrate that the algorithm can run in real-time. In future work, we intend to improve our real-world experiments. In particular, we are looking into implementing a multi-agent solution to account for limited battery-life of UAV's.

\bibliographystyle{IEEEtran}
\bibliography{main.bib}

\end{document}